\documentclass[10pt,twocolumn,letterpaper]{article}
\usepackage{iccv}
\usepackage{times}
\usepackage{epsfig}
\usepackage{graphicx}
\usepackage{amsmath}
\usepackage{amssymb}

\usepackage{booktabs}
\usepackage{url}
\usepackage{multirow}
\usepackage{caption}
\usepackage{subcaption}
\usepackage{pifont}
\usepackage{algorithm}
\usepackage{array}
\usepackage[noend]{algpseudocode}

\usepackage[pagebackref=true,breaklinks=true,letterpaper=true,colorlinks,bookmarks=false]{hyperref}

\iccvfinalcopy 

\def\iccvPaperID{9974} 

\ificcvfinal\fi

\begin{document}

\title{L-DAWA: Layer-wise Divergence Aware Weight Aggregation in Federated Self-Supervised Visual Representation Learning}

\author{Yasar Abbas Ur Rehman\textsuperscript{1,}\thanks{Equal contribution, authors ordered alphabetically.}, 
\space Yan Gao\textsuperscript{2,*}, 
\space Pedro Porto Buarque de Gusm\~{a}o\textsuperscript{2}, 
\space Mina Alibeigi\textsuperscript{2,3}, \\
\space Jiajun Shen\textsuperscript{1}, 
\space Nicholas D. Lane\textsuperscript{2,4}  \\
\small \textsuperscript{1}TCL AI Lab, Hong Kong, \textsuperscript{2}University of Cambridge, United Kingdom, \textsuperscript{3}Zenseact, Sweden, \textsuperscript{4}Flower Labs\\
}

\maketitle
\newcommand{\cmark}{\ding{51}}
\newcommand{\xmark}{\ding{55}}
\newcommand{\uarr}{\uparrow}

\ificcvfinal\thispagestyle{empty}\fi

\begin{abstract}
   The ubiquity of camera-enabled devices has led to large amounts of unlabeled image data being produced at the edge. The integration of self-supervised learning (SSL) and federated learning (FL) into one coherent system can potentially offer data privacy guarantees while also advancing the quality and robustness of the learned visual representations without needing to move data around. However, client bias and divergence during FL aggregation caused by data heterogeneity limits the performance of learned visual representations on downstream tasks. In this paper, we propose a new aggregation strategy termed Layer-wise Divergence Aware Weight Aggregation (L-DAWA) to mitigate the influence of client bias and divergence during FL aggregation. The proposed method aggregates weights at the layer-level according to the measure of \textit{angular divergence} between the clients' model and the global model. Extensive experiments with cross-silo and cross-device settings on CIFAR-10/100 and Tiny ImageNet datasets demonstrate that our methods are effective and obtain new SOTA performance on both contrastive and non-contrastive SSL approaches.
\end{abstract}

\section{Introduction}

Federated Learning (FL) has been a center of interest for the research and industrial communities due to its unique property of collaboratively learning feature representations from large-scale datasets without compromising the users' data privacy \cite{mcmahan2017communication, zhao2018federated,li2020review,kairouz2021advances}.
It has been successful in joint visual representations learning from image data while preserving data privacy \cite{mcmahan2017communication,reddi2020adaptive,li2020federated}. 
However, current practices in FL are commonly limited to supervised learning tasks that require high-quality and domain-specific labels to be available alongside the data. This requirement limits the deployment of FL in many real-world applications where access to high-quality labels at the edge is restricted \cite{jain2021biometrics}. 

Self-Supervised Learning (SSL) has been combined with FL, enabling it to expand its potential of learning feature representations from the vast amount of unlabeled, private, uncurated, and visual data being produced at the edge \cite{zhuang2021collaborative, lubana2022orchestra, li2021model, zhuang2022divergence}. 
In contrast to supervised FL \cite{dong2022spherefed, gao2022end, mcmahan2017communication}, federated self-supervised learning (F-SSL) does not require high-quality labeled data, although it may require another stage of centralized fine-tuning or personalizing the model for downstream tasks with limited labeled data. F-SSL enables the re-purposing of heterogeneous, unlabeled, and uncurated real-world image data for various downstream tasks. (\textit{viz}., image recognition \cite{krizhevsky2009learning}, object detection\cite{he2016deep}, semantic segmentation\cite{park2020sinet}, facial recognition, authentication \cite{jain2021biometrics, schroff2015facenet}, etc.) by collaboratively learning \textit{intermediate} visual representations in a privacy-preserving fashion. 

\begin{figure*}
    \centering
    \includegraphics[width=0.95\linewidth]{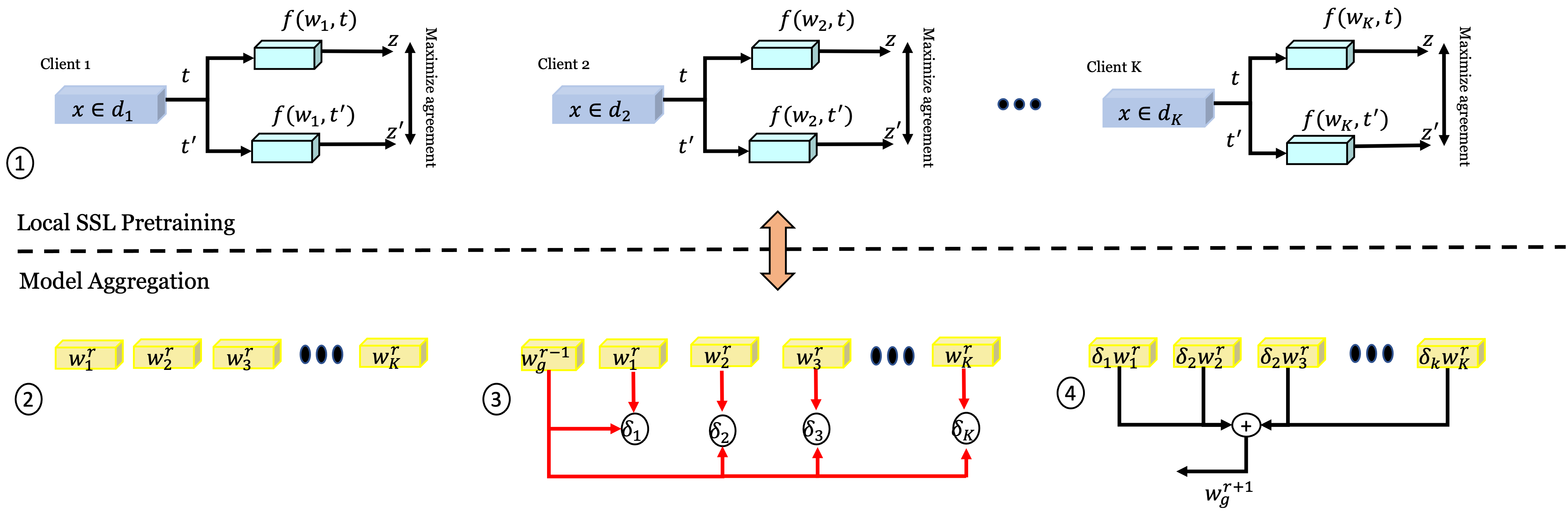}
    \caption{\small Pipeline of DAWA for Federated SSL. \textcircled{1} Local SSL pre-training. \textcircled{2} Received clients' models on the server. \textcircled{3} Computing layer-wise divergence for each client model. \textcircled{4} Aggregating the clients' models weighted by the corresponding layer-wise divergence and generating a new global model. Note that the client models are discarded after generating the global model.} 
    \label{fig:model}
    \vspace{-3mm}
\end{figure*}

One of the unique challenges of F-SSL is learning visual representations from non-independently and identically distributed (\textit{Non-iid}) data \cite{zhuang2021collaborative, lubana2022orchestra, rehman2022federated}. Recent studies in F-SSL, both image-based \cite{zhuang2021collaborative} and video-based \cite{rehman2022federated}, directly extend the state-of-the-art (SOTA) centralized SSL pre-training techniques (e.g., SimCLR \cite{chen2020simple}, SimSiam \cite{chen2021exploring}, BYOL \cite{grill2020bootstrap}, Barlow Twins \cite{zbontar2021barlow} Speed \cite{Benaim_2020_CVPR}, VCOP \cite{xu2019self}, and CtP \cite{wang2021unsupervised}) under the setting of FL. Generally, these F-SSL methods aggregate the participating clients' model during each FL round using FedAvg \cite{mcmahan2017communication}. While FedAvg provides certain convergence guarantees during the FL stage, the downstream task performance is sub-optimal \cite{caldarola2022improving}. 

One of the main reasons for sub-optimal performance is the uncontrolled divergence between participating clients' caused by data heterogeneity. \cite{zhao2018federated, li2019fair, zhuang2021collaborative, zhuang2022divergence}. 
The client's model divergence in FL, if not appropriately controlled, affects the aggregated global model, even if the data and the model quality of individual clients are good, which often happens when the clients' models lie in different basins of the loss landscape \cite{wortsman2022model, li2022understanding}.  
FedAvg, used by most existing F-SSL work, aggregates the local models based on the number of data samples on each client. This leads the model to be biased towards the clients possessing more data \cite{li2019fair}, and the optimization trajectory of the global model to be dominated by those clients, thus deviating the model into local minima, further leading to sub-optimal downstream performance. While bias is common in both \textit{cross-silo} and \textit{cross-device} settings of FL, it can have a more significant influence in the former, especially when all clients participate in each FL round to generate the global model and clients with more data dominate the training. In the \textit{cross-device} settings, a fraction of clients participate in every round of FL, mitigating the possibility of certain groups of clients dominating the FL training; however, client drift is higher compared to \textit{cross-silo}, resulting in a worse global model when quality clients are not selected. 


To alleviate this issue, several aggregation strategies based on model divergence among clients have been proposed.  FedU~\cite{zhuang2021collaborative} determines the update of the local client models' predictor based on the divergence between the backbone models of the clients and the server. Despite its effectiveness, FedU still uses FedAvg~\cite{mcmahan2017communication} to aggregate client models at the server. This overlooks the problem of clients' bias caused by FedAvg during model aggregation. 
Another work termed Loss \cite{gao2022end} uses clients' local training loss as an indicator of the clients' model quality to aggregate models. However, this method is still biased toward the clients with lower training loss (similar problem as FedAvg).
Additionally, the existing aggregation strategies assign a single scalar to one client model. Nevertheless, the different layers of the clients' model exhibit different levels of heterogeneity \cite{Ma_2022_CVPR, zhang2021parameterized}. Assigning a single scalar value to all layers of a client's model would exacerbate the bias.



\begin{figure*}
\centering
\begin{subfigure}{0.25\linewidth}
    \includegraphics[scale=0.4]{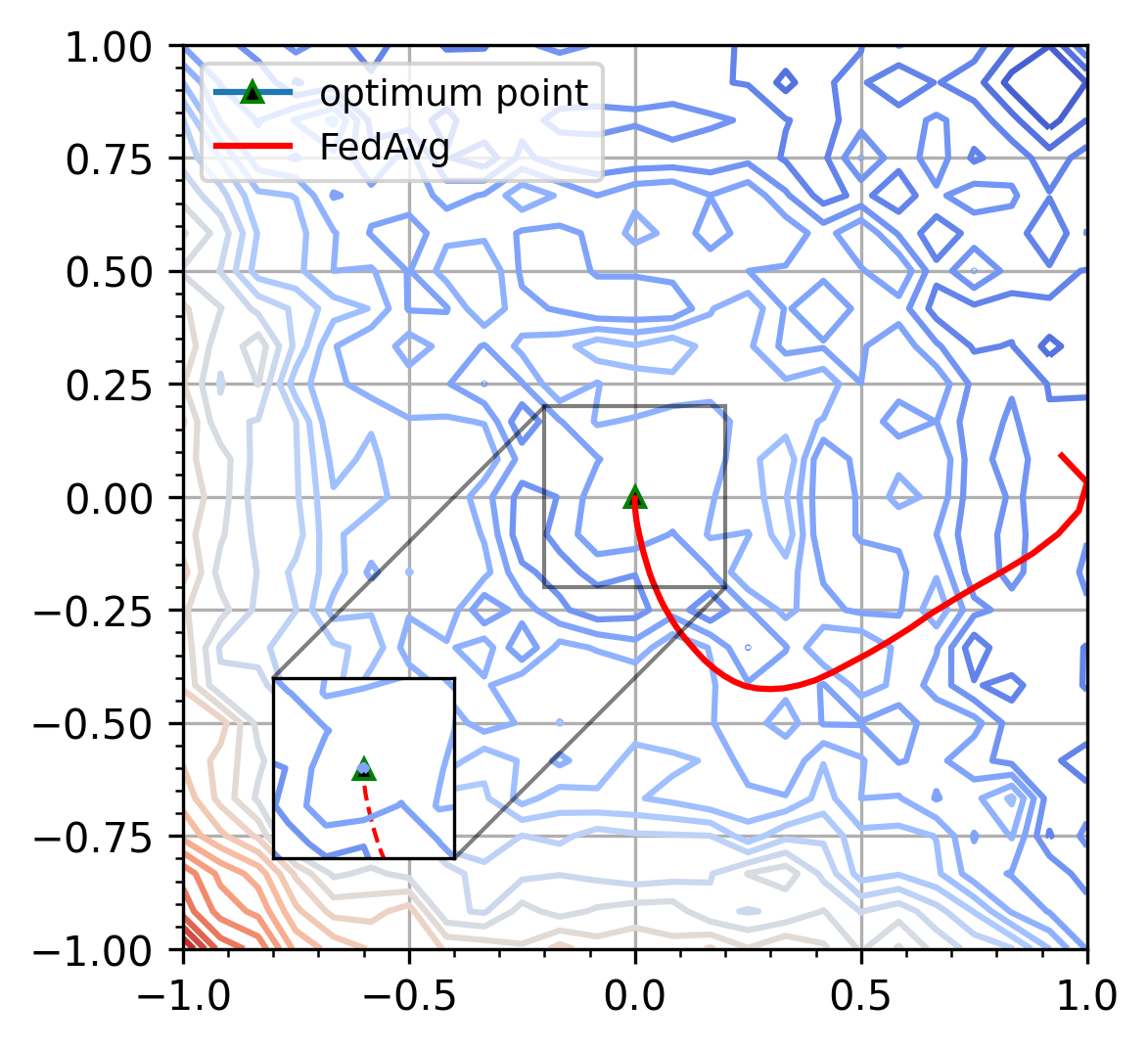}
    \subcaption{\scriptsize FedAvg SimCLR}
\end{subfigure}%
\begin{subfigure}{0.25\linewidth}
    \includegraphics[scale=0.4]{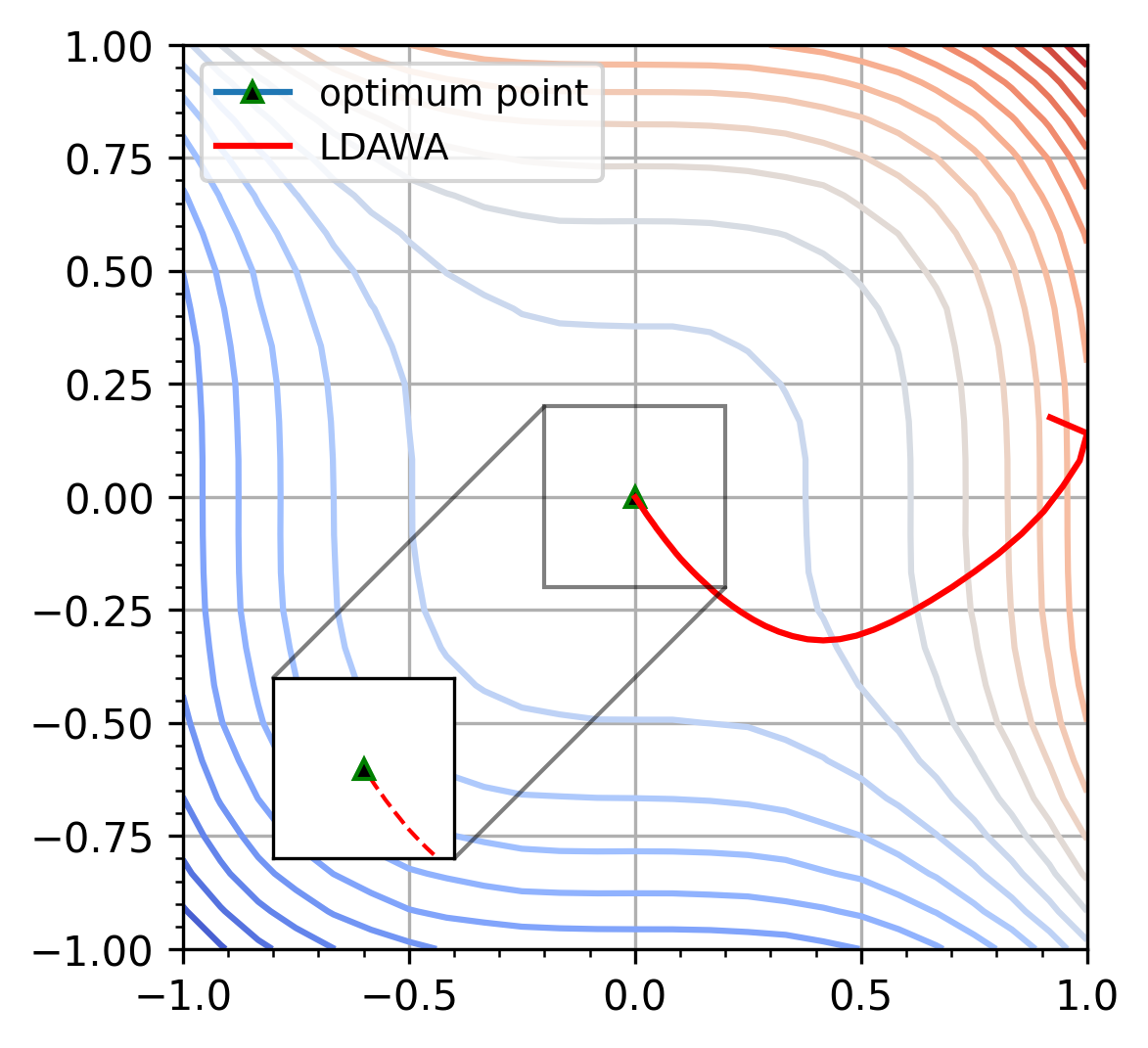}
    \subcaption{\scriptsize L-DAWA SimCLR}
\end{subfigure}%
\begin{subfigure}{0.25\linewidth}
    \includegraphics[scale=0.4]{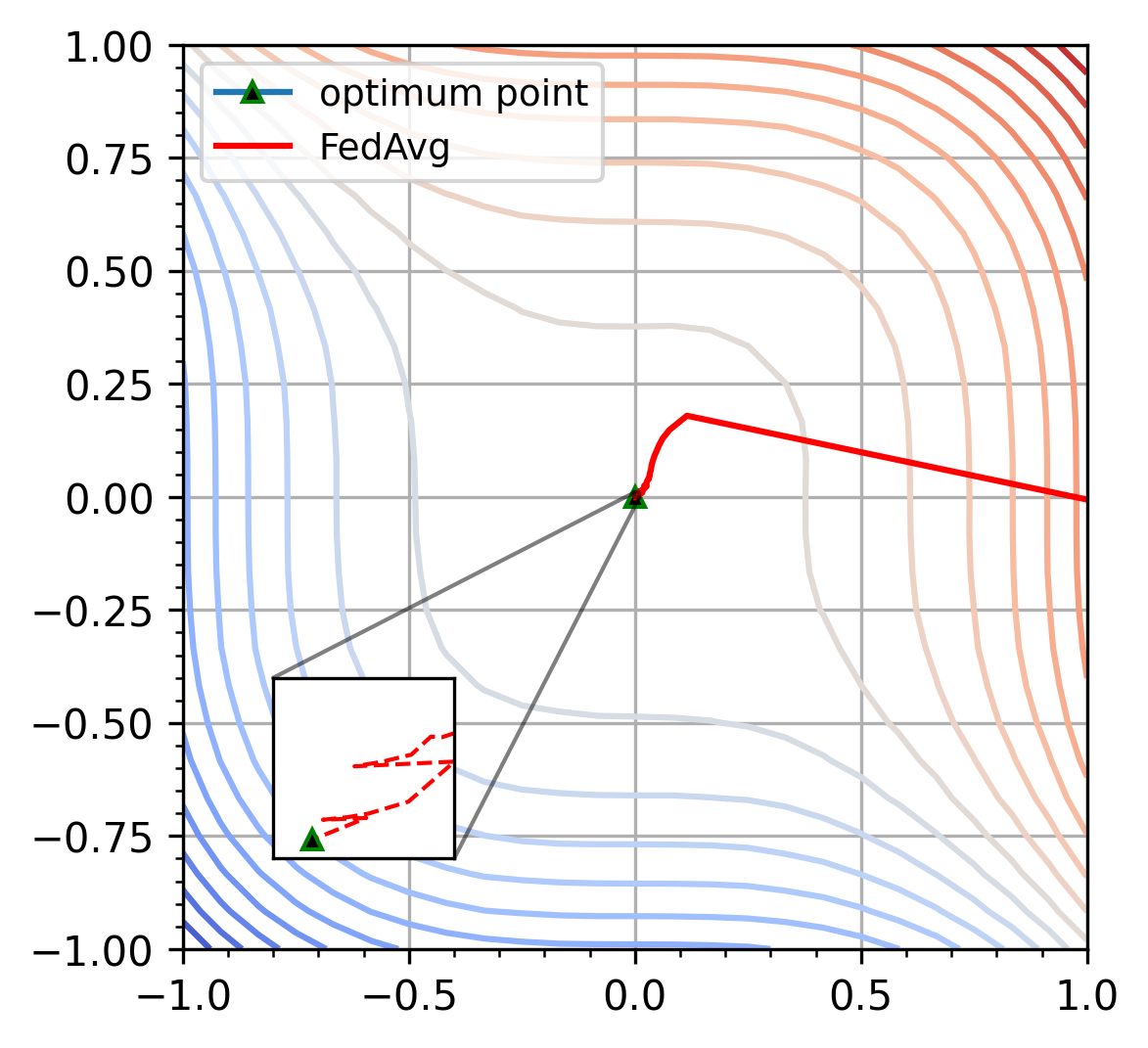}
    \subcaption{\scriptsize FedAvg Barlow Twins}
\end{subfigure}%
\begin{subfigure}{0.25\linewidth}
    \includegraphics[scale=0.4]{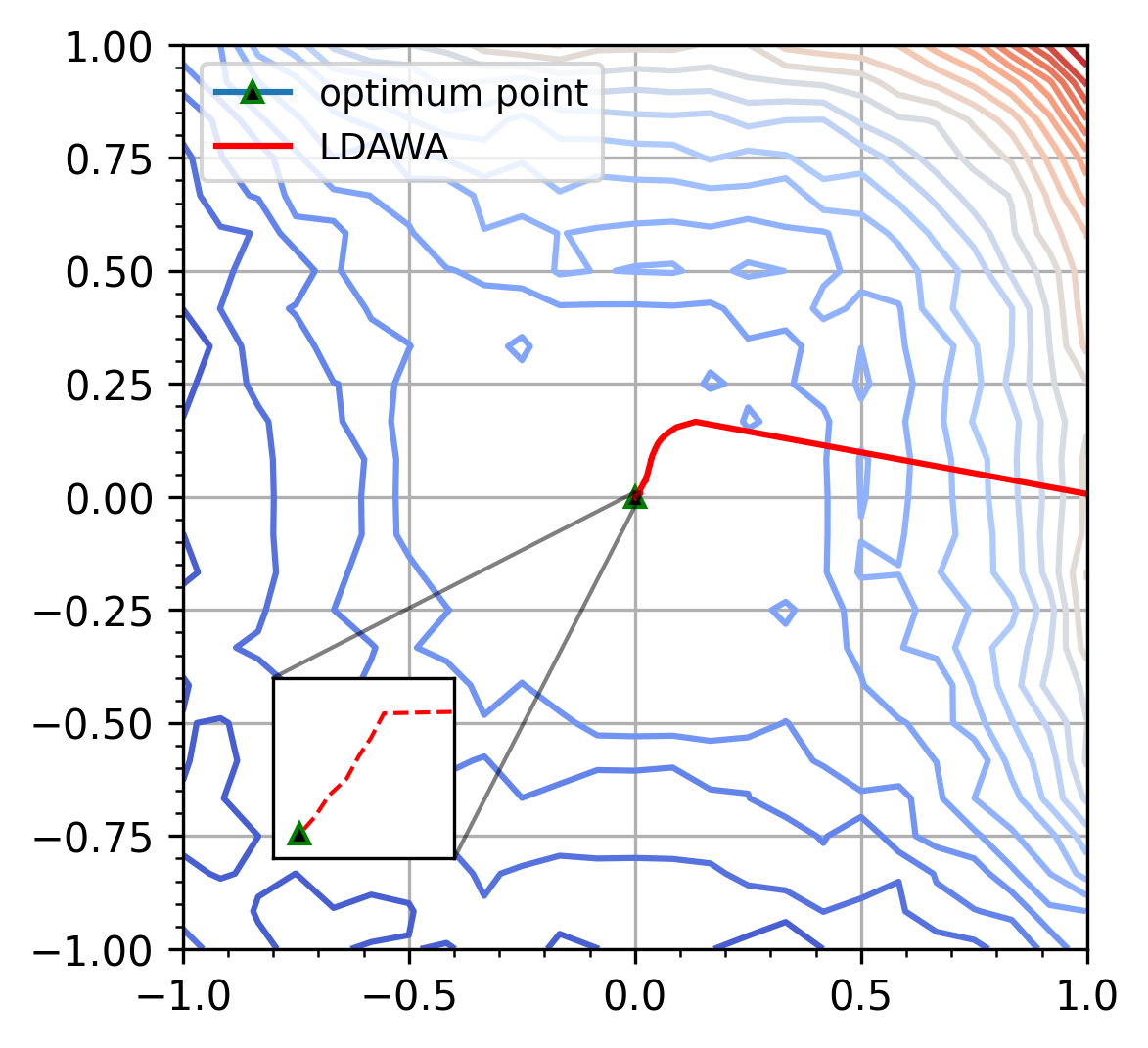}
    \subcaption{\scriptsize L-DAWA Barlow Twins}
\end{subfigure}
    
    \caption{\small Illustration of global model optimization trajectories for FedAvg \cite{mcmahan2017communication} and L-DAWA under the loss-landscapes of FL SimCLR (a-b) and FL Barlow Twins (c-d) with \textit{cross-silo} settings. 
    L-DAWA assists the global model in converging smoothly into a substantially wider loss landscape compared to FedAvg (especially for FL SimCLR with a chaotic loss landscape in (a)). The optimization trajectories are effectively controlled by L-DAWA. For instance, (b) achieves a shorter path toward the optimum point than (a); (d) obtains a more smooth converge route near the optimum point compared to (c), which suffers clear oscillations.}
    
    
    \label{fig:Optimization-a}
    \vspace{-3mm}
\end{figure*}

In this paper, we propose a novel aggregation scheme termed Layer-wise Divergence Aware Weight Aggregation (L-DAWA) to mitigate the dominant effect of certain clients in F-SSL. L-DAWA is a unique method that incorporates \textit{angular divergence} at the layer level into the aggregation process. The \textit{angular divergence} is used as a weighting coefficient to scale the contribution of each layer from the client's models in the generation of the global model at the server. 
L-DAWA does not require the transmission of any clients' metadata, such as the number of samples and local training loss, to the server. Instead, it relies solely on the previous global model and the local clients' model updates. By integrating angular divergence during the aggregation process, similar to the role of momentum, it is possible to achieve smooth control of the trajectory of model optimization and accelerate convergence in the relevant direction. This would lead the model toward the global optima, resulting in better downstream performance (Figure \ref{fig:Optimization-a}, Table \ref{tab:comp_sota_cross-silo-a}).


We further adapt L-DAWA into the existing aggregation methods (FedAvg, Loss, and FedU) to correct the optimization trajectory dominated by certain clients. We termed these modified techniques as L-DAWA$_{FedAvg}$, L-DAWA$_{FedU}$, and L-DAWA$_{Loss}$. Experimental results show that the proposed method and its variants significantly improve the performance both in \textit{cross-silo} and \textit{cross-device} settings.
The main contributions of this work are as follows:
\begin{itemize}
    \item We integrate for the first time the angular divergence into model aggregation, which maintains a coherent convergence trajectory during FL training, leading to better models, see Table~\ref{tab:comp_sota_cross-silo-a}.
    \vspace{-1mm}
    \item We propose a novel aggregation method for F-SSL, dubbed L-DAWA, which utilizes angular divergence as a weighting coefficient to scale the contribution of each layer from the client's models during FL aggregation. We further adapt L-DAWA into the existing aggregation methods (FedAvg, Loss, and FedU).
    \vspace{-1mm}
    
    \item  We perform extensive experiments and compare the performance of L-DAWA-related methods for both contrastive and non-contrastive SSL approaches. We achieve the new SOTA in F-SSL obtaining at most $5.21$\%, $6.19$\%, and $6.25$\% improvement on CIFAR-10/100 and Tiny ImageNet, respectively under linear evaluation protocol in \textit{cross-silo} and comparable performance in \textit{cross-device} setting. 
\end{itemize}




\section{Literature review}
\subsection{Self-supervised learning}
Self-supervised learning has been explored predominantly in every computer vision domain \cite{goyal2022vision}. Owing to learning representations based on exploiting the properties in the data using some generative or discriminative models that solve pseudo or pretext tasks. Examples of generative self-supervised pretext tasks include colorization \cite{zhang2016colorful}, in-painting \cite{pathak2016context}, or super-resolution \cite{Menon_2020_CVPR}. On the other hand, discriminative self-supervised pretext tasks include, but are not limited to, predicting rotation \cite{komodakis2018unsupervised}, feature alignment \cite{chen2020simple, he2020momentum,grill2020bootstrap,caron2020unsupervised, zbontar2021barlow, yang2021instance}, solving jigsaw puzzle (a.k.a. predicting shuffled patch permutation) \cite{noroozi2016unsupervised}. Among these discriminative SSL approaches, feature alignment approaches such as SimCLR \cite{chen2020simple}, MOCO \cite{he2020momentum}, BYOL \cite{grill2020bootstrap}, SWAV \cite{caron2020unsupervised}, and Barlow Twins \cite{zbontar2021barlow} have been in the spotlight recently. Based on their loss function, these SSL approaches can be categorized into contrastive (requires negative samples), and non-contrastive (does not require negative samples) approaches. They can be combined into a single category of feature alignment because they enable the model to learn features that are invariant to the artificial transformation generated via the data augmentation process. 

\begin{figure*}
\centering
\begin{subfigure}{0.5\linewidth}
    \centering
    \includegraphics[width=\linewidth]{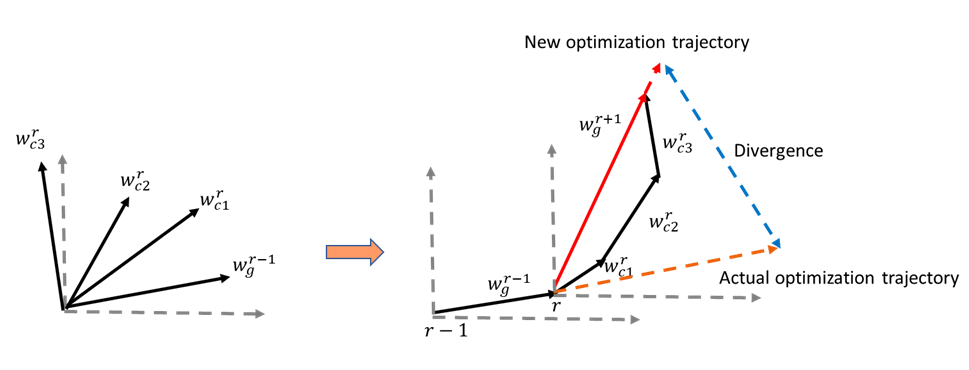}
    \caption{FedAvg}
\end{subfigure}%
\begin{subfigure}{0.5\linewidth}
    \centering
    \includegraphics[width=\linewidth]{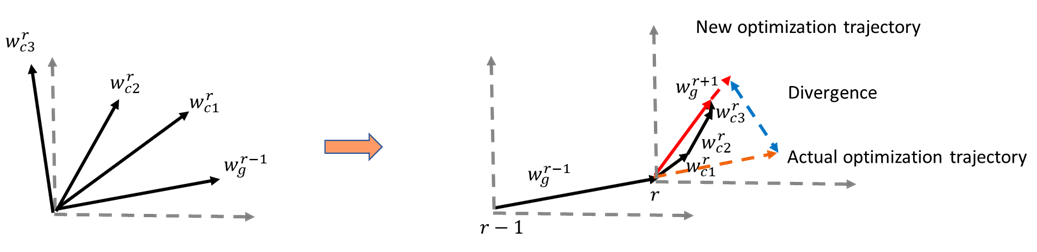}
    \caption{DAWA}
\end{subfigure}

    \caption{\small Illustration of optimization trajectories for FedAvg and DAWA. (a) Without any divergence control, FedAvg can deviate from the actual optimization trajectory by a large margin 
    (b) DAWA effectively controls the trajectory by scaling the clients based on the measure of their \textit{angular divergence }compared to the previous global model, effectively restricting the \textit{angular divergence} range of the clients.} 
    \label{fig:Optimization-b}
    \vspace{-3mm}
\end{figure*}


\subsection{Data heterogeneity in FL}
Data heterogeneity is inevitable in realistic FL settings and has been studied extensively in literature \cite{mcmahan2017communication,zhao2018federated,lubana2022orchestra,li2021model, wang2022friends, zhang2020federated}.  McMahan et al. \cite{mcmahan2017communication} proposed a client models aggregation strategy, FedAvg, that can work on certain Non-iid distributions. FedAvg has been considered as a baseline for many FL aggregation strategies that improve the performance of the model in Non-iid data settings \cite{gao2022end, reddi2020adaptive}. Data heterogeneity in FL causes weight divergence, i.e., during the aggregation, the weights of the model from different clients are not aligned causing certain weights to cancel each other \cite{Mendieta_2022_CVPR}. Methods have been proposed to tackle the weight divergence in F-SSL, such as by introducing global dataset \cite{zhao2018federated}, local training loss \cite{gao2022end}, or locally updating the partial model based on the measure of divergence \cite{zhuang2021collaborative, zhuang2022divergence}.     

\subsection{Divergence in F-SSL}
Studies have been conducted to understand the clients' model divergence in F-SSL \cite{zhuang2021collaborative, zhuang2022divergence}. 
These frameworks allow to partially update the local SSL model (e.g., only updating the predictor) based on the Euclidean distance between the current model and previous global model.
Although these methods are effective on the downstream task, they are limited to the only non-contrastive SSL technique, e.g., BYOL\cite{grill2020bootstrap}. Moreover, these methods naively combine the clients' models at the server (using FedAvg) and push the computation of the model divergence on the clients' side, requiring the clients to perform extra computation.


Using angular distance for measuring the divergence or similarity between the weight vectors in supervised \cite{ou2020lincos} and unsupervised FL is not a new concept. Several works use angular distance for client clustering.
For example,  \cite{sattler2020clustered} computes the cosine distance to cluster different clients for multi-task learning after fetching the learned global model. In the other works, the cosine-based similarity is used to perform cluster-based supervised FL of different clients \cite{duan2021flexible}, or to understand the divergence between the representations of different clients' predictor networks \cite{dong2022spherefed}. 
The work in \cite{Ma_2022_CVPR} uses hypernetworks to find the similarity between the weights of the clients' models for personalized FL. However, the personalized models for each client are maintained on the server, which differs from a general FL framework, where no client's models are reserved on the server.
We further note that, to the best of our knowledge, there is no existing work considering angular divergence for model aggregation in the general FL framework, which is considered in our work. 


\section{Methodology}

\subsection{Federated self-supervised learning}
We consider $M$ partitions $\{d_m\}_{m=1}^{M}$ of dataset $D$ to compose $M$ decentralized clients with $\{n_m\}_{m=1}^{M}$ samples on each local data set. At each communication round $r$ of FL, the server randomly selects $K$ clients participating in the training and initializes the local models with the global model weights $w_{g}^{r}$. Then, each decentralized client $k$ learns the intermediate feature representations $w_{k}^r=\mathcal{F}_{SSL}(d_{k}, w_{g}^{r}, E)$ by training with a specific SSL approach on its own local dataset $d_{k}$ for $E$ local epochs before transmitting the local model $w_{k}^{r}$ to the server. The server then receives the local models $\{w_{k}^{r}\}_{k=1}^{K}$ and aggregate them based on a weighting factor $\beta(\cdot)$ to generate a new global model $w_{g}^{r+1}$ as follows:

\begin{equation}
    w^{r+1}_{g} = \sum_{k=1}^{K}\beta_{k}w_{k}^{r}.
    \label{eq:1}
\end{equation}

This process is repeated until model convergence. A common aggregation strategy is FedAvg \cite{mcmahan2017communication}, which combines the local models based on the number of samples over selected clients, i.e., $\beta_{k} = \frac{n_k}{\sum_{k=1}^K n_k}$. In realistic FL settings with heterogeneous client data distribution, some clients may contain skewed data, not representing the global data distribution. This scenario could lead to model deviation in the aggregation step, which can not be solved by vanilla FedAvg. 
Several existing works \cite{zhuang2021collaborative,lee2021layer} attempt to alleviate this issue by using Euclidean distance (between the client's model and the global model) as an indicator to partially or entirely update the layers of the client models. Nevertheless, the aggregation step on the server for these methods is still based on FedAvg. Other works \cite{gao2022end,rehman2022federated} use the averaged training loss as a weighting coefficient for aggregation, i.e., $\beta_{k} = \frac{exp(-\mathcal{L}_{k})}{\sum_{k=1}^{K}exp(-\mathcal{L}_{k})}$, thus reflecting the quality of the locally trained models. 


However, the above approaches are generally biased toward the clients with a larger number of data samples or lower training loss. 
In this case, the trajectory of the global model optimization is determined by certain clients \cite{li2019fair}, thus deviating into some local minima \cite{zhao2018federated,liu2020accelerating} resulting in sub-optimal performance (Figure \ref{fig:Optimization-a}).
On the model level, the dominant effect of certain clients during FL training leads to oscillations in the divergence of the clients' models with respect to the global model (Figure \ref{fig:div_acc}). This could deviate the model optimization trajectory, which results in a decrease in performance on the downstream tasks (see Table \ref{tab:comp_sota_cross-silo-a}).


\subsection{Divergence aware weight aggregation}
A smooth control of the trajectory of model optimization would assist model convergence and leading to higher downstream performance. Here, we propose a divergence aware weight aggregation (DAWA) method by introducing the \textit{angular divergence} $\delta_{k}$ between the clients' model and the previous global model into aggregation process (Figure \ref{fig:model}). $\delta_{k}$ can be calculated as follows:

\begin{equation}
 \delta_{k} = cos\theta_{g,k} = \frac{w_{g}^{r} . w_{k}^{r}} { ||w_{g}^{r}||. ||w_{k}^{r}||}.
 \label{eq:cosine}
\end{equation}

The angular divergence determines whether the model weights are aligned or orthogonal to each other. The range of $\delta_{k}$ is naturally restricted to [-1, 1].
To integrate angular divergence into aggregation, we set $\beta_k = \delta_{k} / K$, then Eq. \ref{eq:1} can be re-written as:

 \begin{equation}
w^{r+1(M-DAWA)}_{g} =  \frac{1}{K}\sum_{k=1}^{K} \delta_{k} w_{k}^{r}. 
 \label{eq:M-DAWA}
\end{equation}

\begin{figure}
\begin{subfigure}{0.32\linewidth}
\includegraphics[scale=0.62]{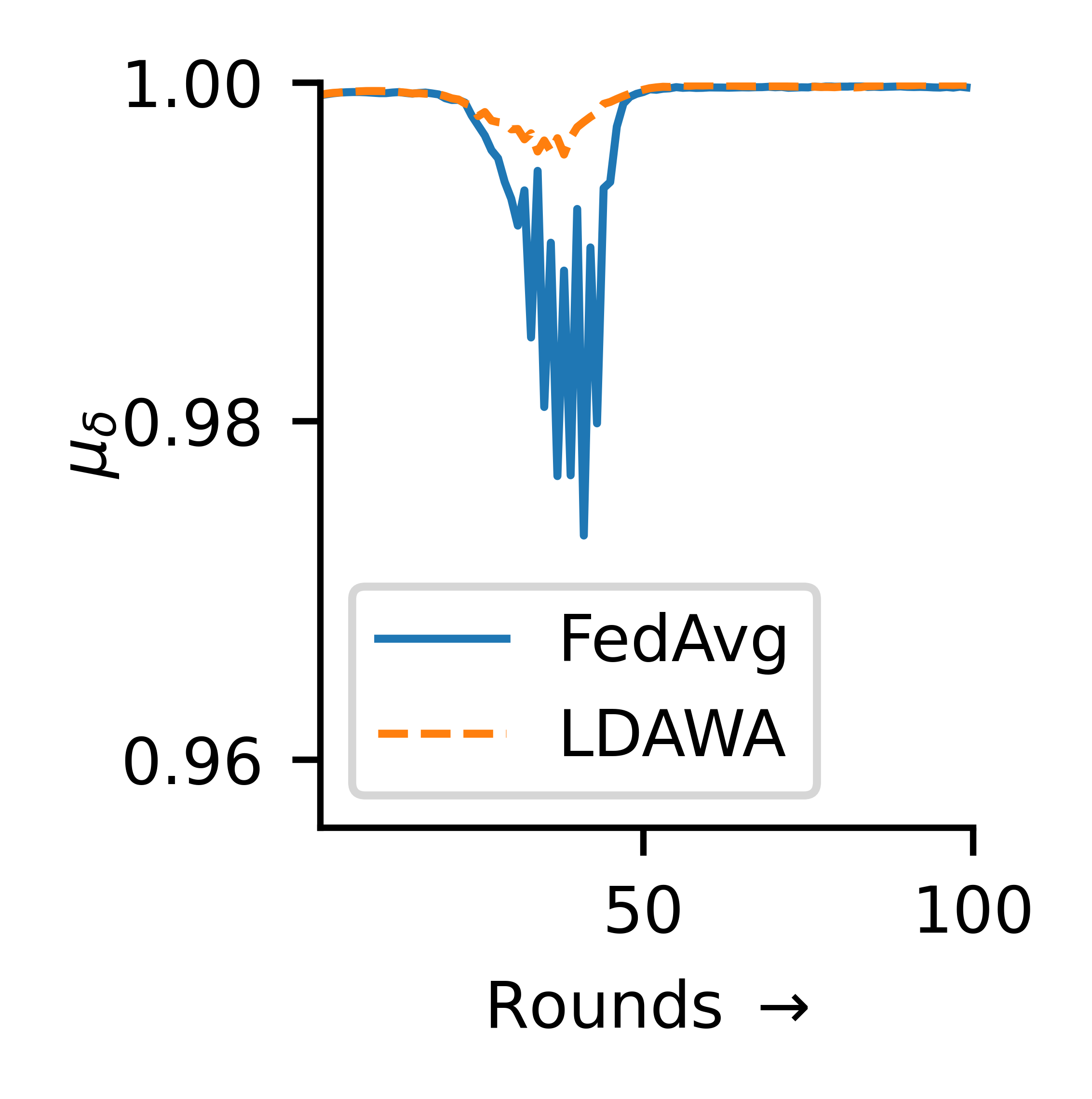}
\subcaption{E=1}
\end{subfigure}%
\hfill
\begin{subfigure}{0.32\linewidth}
    \includegraphics[scale=0.62]{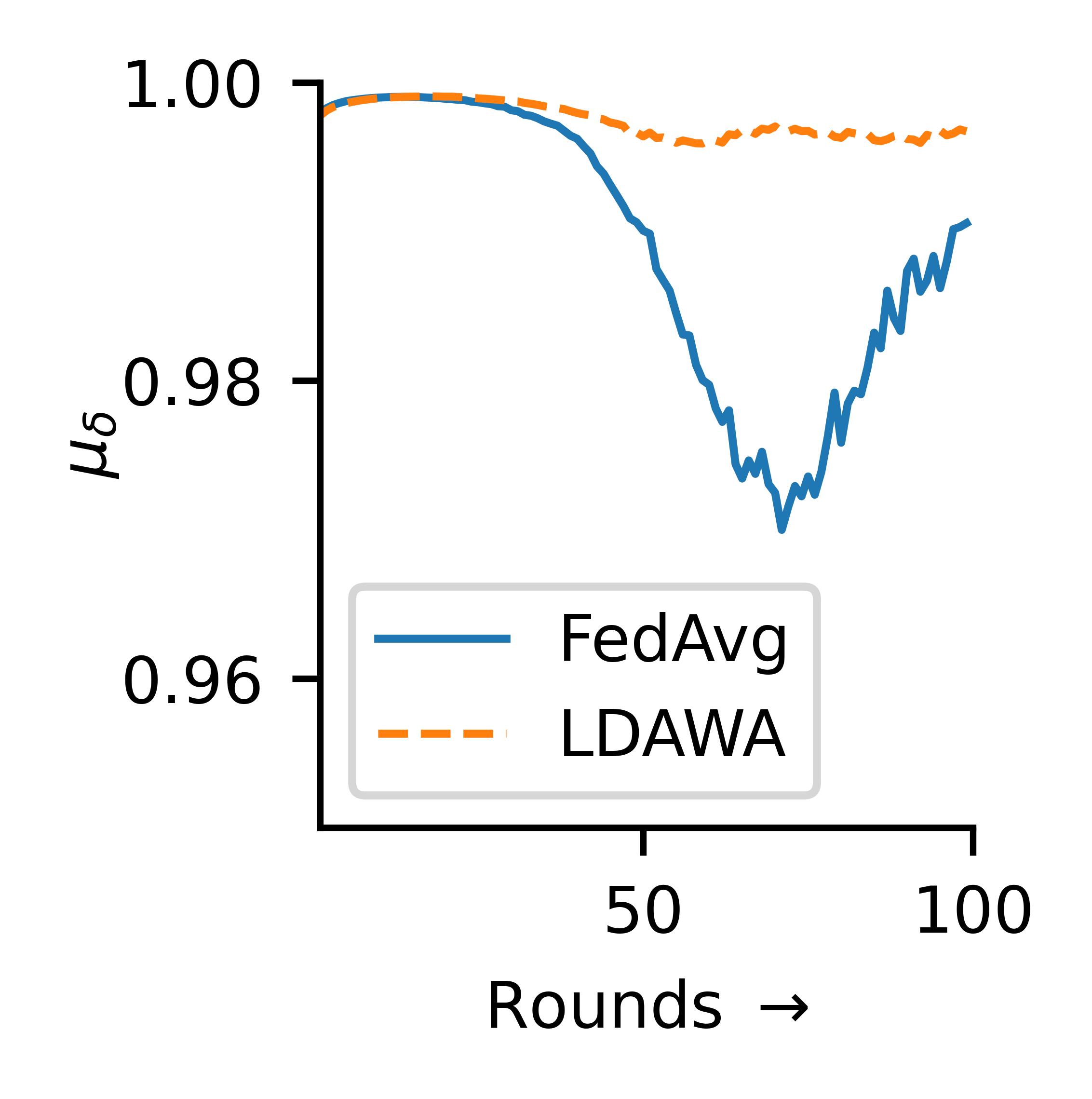}
    \subcaption{E=5}
\end{subfigure}%
\hfill
\begin{subfigure}{0.32\linewidth}
 \includegraphics[scale=0.62]{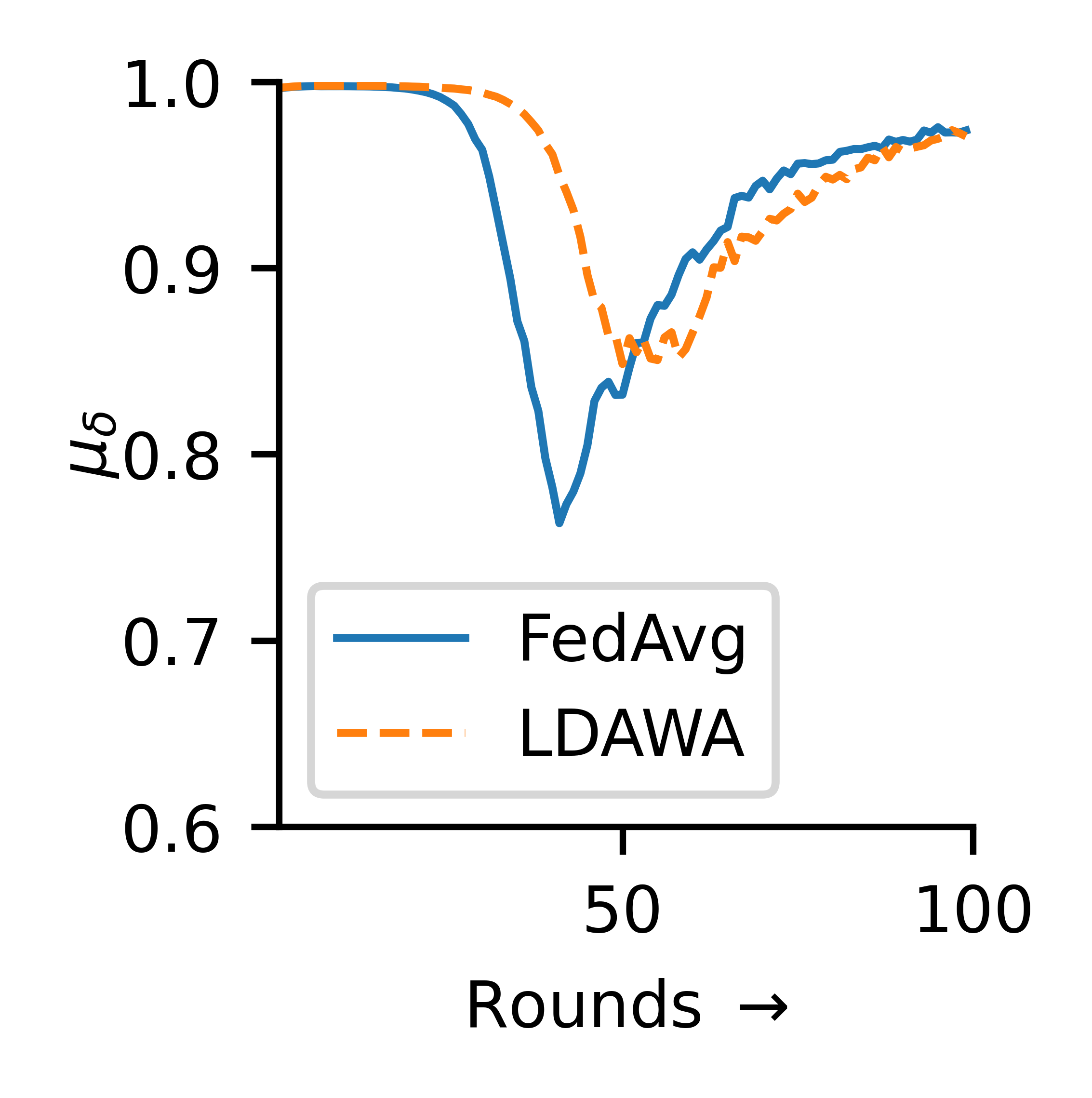}
 \subcaption{E=10}
\end{subfigure}%
\hfill
    \caption{\small The mean angular divergence between the clients' models with respect to the previous global model over $R=100$ rounds, computed by the following equation: $\mu_{\delta} = \frac{1}{K} \sum_{k=1}^{K}\delta_{k}$, where $K=10$. The higher $\delta$ value means lower divergence. L-DAWA has a good control of oscillations,  maintaining the angular divergence in a lower level over FL rounds than FedAvg.}
    \label{fig:div_acc}
    \vspace{-3mm}
\end{figure}

The integration of angular divergence, as a similar role of momentum, would accelerate convergence in the relevant direction while dampening oscillations during FL optimization (Figure \ref{fig:Optimization-a} \& \ref{fig:Optimization-b} \& \ref{fig:div_acc}).
For instance, some client models $w_{k}^{r}$ may diverge by a large angle (e.g., $[180^{\circ}, 90^{\circ})$) from the global model $w_{g}^{r}$. In this case, $\delta_{k}$ in Eq. \ref{eq:cosine} falls into the range of $[-1,0)$. The subsequent multiplication of $\delta_{k}$ with $w_{k}^{r}$ in Eq. \ref{eq:M-DAWA} could correct the alignment with the direction of $w_{g}^{r}$ by effectively reducing the theoretical angular range of divergence from $[180^{\circ}, 0^{\circ}]$ to $[90^{\circ}, 0^{\circ}]$. 


However, the divergence at layer-level may dramatically vary, which cannot be represented by a single value $\delta_{k}$ of the whole model. Additionally, directly computing $\delta_{k}$ on the entire model is prohibitively expensive in practice imposing large memory footprints and increase the computation time (see Table \ref{tab:DAWA-P-vs-DAWA-C}). As a result, we further calculate angular divergence $\delta_{k}^{(l)}$ based on Eq. \ref{eq:cosine} for each layer. Then, the layer-wise divergence aware weight aggregation (L-DAWA) method can be represented as follows:

\begin{equation}
w^{r+1(L-DAWA)}_{g} =  \frac{1}{K}\sum_{l=1}^{L}\sum_{k=1}^{K}  \delta_{k}^{(l)} w_{k}^{r(l)}. 
 \label{eq:L-DAWA}
\end{equation}

\begin{algorithm}[t]
            \caption{\small \textit{L-DAWA}: Let us consider the server randomly selecting $K$ clients at the given round. The clients train the SSL model with $L$ layers for $E$ local epochs on its dataset $d_k$ with $n_k$ number of samples. The FL optimization lasts $R$ rounds.}
            \label{al1}
            \textbf{Input}: $K, R, n_k, d_k, E, L$ \\
            \textbf{Output}: $w_{g}^R$ \\
            \textbf{Central server does:}
            \begin{algorithmic}[1]
            \For{$r = 1,$...$,R$}
                \State Server randomly selects $K$ clients.
                \For{$k = 1,$...$,K$}
                    \State $w_{k}^{r}, n_{k}, \mathcal{L}_{k}$ =  \textbf{TrainLocally}$(k, w_{g}^r, E)$
                \EndFor
                
                \State \textbf{Aggregation}:
                \For{$l = 1,$...$,L$}
                    \State \small Compute divergence $\delta_k^{(l)}$ based on Eq. \ref{eq:cosine}.
                \EndFor
                \State \textbf{If \textit{M-DAWA} then}: \small Compute $w_g^{r+1}$ based on Eq. \ref{eq:M-DAWA}.
                \State \textbf{If \textit{L-DAWA} then}: \small Compute $w_g^{r+1}$ based on Eq. \ref{eq:L-DAWA}.
                \State  \textbf{If \textit{L-DAWA$_{FedAvg}$} or} \textbf{\textit{L-DAWA$_{FedU}$} then}:
                \State \small ~~~~ Compute $w_g^{r+1}$ based on Eq. \ref{eq:L-DAWA-fedavg}. 
                \State \textbf{If \textit{L-DAWA$_{Loss}$} then}:
                \State \small ~~~~ Compute $w_g^{r+1}$ based on Eq. \ref{eq:L-DAWA-loss}. 
            \EndFor 
            \end{algorithmic}
            
            \textbf{TrainLocally $(k,w_g^r$):}
            \begin{algorithmic}[1]
            \State  $w_{k}^r, \mathcal{L}_{k} =\mathcal{F}_{SSL}(d_{k}, w_{g}^{r}, E)$
            \State  Upload $w_{k}^r, n_{k}, \mathcal{L}_{k}$ to the server.
            \end{algorithmic}
\end{algorithm}

\subsection{Layer-wise divergence adaptation with SOTA aggregation strategies}
The layer-wise angular divergence can be easily integrated to the existing aggregation methods in order to correct the trajectory of model optimization. Here, we select three SOTA aggregation strategies, namely FedAvg \cite{mcmahan2017communication}, Loss \cite{gao2022end}, and FedU \cite{zhuang2021collaborative}. By introducing divergence, we term the new variations as L-DAWA$_{FedAvg}$ (Eq. \ref{eq:L-DAWA-fedavg}), L-DAWA$_{Loss}$ (Eq. \ref{eq:L-DAWA-loss}) and L-DAWA$_{FedU}$ (Eq. \ref{eq:L-DAWA-fedavg}). Note that FedU conducts aggregation with FedAvg, and partially update the model based on Euclidean distance. The overall algorithm is summarised in Algo. \ref{al1}.

\begin{equation}
\small w^{r+1(FedAvg)}_{g} =\sum_{l=1}^{L} \sum_{k=1}^{K}  \frac{n_k}{\sum_{k=1}^K n_k} \delta_{k}^{(l)} w_{k}^{r(l)}. 
 \label{eq:L-DAWA-fedavg}
\end{equation}

\begin{equation}
\small w^{r+1(Loss)}_{g} = \sum_{l=1}^{L} \sum_{k=1}^{K} \frac{exp(-\mathcal{L}_{k})}{\sum_{k=1}^{K}exp(-\mathcal{L}_{k})} \delta_{k}^{(l)} w_{k}^{r(l)}. 
 \label{eq:L-DAWA-loss}
\end{equation}

\section{Experimental setup}
\subsection{Federated datasets}
\label{sec:dataset}
For both F-SSL pre-training and downstream tasks evaluation, we conduct experiments on CIFAR-10/100 and Tiny ImageNet datasets \cite{krizhevsky2009learning}. To simulate a realistic FL environment, we generate Non-iid versions of datasets based on actual class labels using a Dirichlet coefficient $\alpha$, where a lower value indicates greater heterogeneity. As a result, the datasets are randomly partitioned into 10/100 shards for \textit{cross-silo}/\textit{cross-device} setting to mimic the setup of having 10/100 disjoint clients participating in FL. Additionally, we scale the value of $\alpha$ to generate various Non-iid levels w.r.t. class labels and the number of samples. 

\subsection{Federated SSL pre-training}
\label{sec:fl_pretraining}
To perform a side-by-side comparison of the contrastive and non-contrastive SSL methods, we select SimCLR \cite{chen2020simple} and Barlow Twins \cite{zbontar2021barlow}. Both SSL approaches share the same network architecture (ResNet-18) \cite{he2016deep, tamkin2021viewmaker} as a backbone while trained with different loss functions. To evaluate these methods on common ground, we maintain the same hyperparameters for both these methods throughout the FL pre-training, except the hyperparameters $\tau$ in SimCLR and $\lambda$ in Barlow Twins. We adopted the setup from Tamkin et al. \cite{tamkin2021viewmaker}, using a batch size of 256, a learning rate of 0.03, a weight decay of 1e-4, and an SGD momentum of 0.9. Unless otherwise stated, we set the number of local epochs $E=10$. For \textit{cross-silo} FL, all clients participate in the training each round, while 10 clients participate per round in \textit{cross-device} setting. The FL pre-training lasts for R=200 rounds for \textit{cross-silo} and R=100 rounds for \textit{cross-device} settings. All training schemes are implemented with PyTorch-Lightning \cite{falcon2019pytorch}, and Flower \cite{beutel2020flower}.



\begin{table}[t]
    \centering
    
    \scalebox{0.8}{
    \begin{tabular}{l cccc}
    \toprule
         Method & Execution Time (Sec) & E=1 & E=5 & E=10\\
         \hline
         FedAvg (Baseline)  & 0.29 &  52.90 & 63.76 & 67.64 \\
         M-DAWA & 22.65& 53.87 & \textbf{65.80}& 68.90\\
         L-DAWA & 0.38 & \textbf{53.94} & 65.44 & \textbf{69.34} \\
         
        
         \bottomrule
    \end{tabular}
    }
    \caption{\small Ablation study: Linear-probe accuracy on downstream task and average aggregation execution time for FedAvg, M-DAWA and L-DAWA. Each method is pre-trained with SimCLR on the Non-iid version ($\alpha$=0.1) of CIFAR-10 for R=10 rounds under the \textit{cross-silo (K=10)} settings.}
    \label{tab:DAWA-P-vs-DAWA-C}
    \vspace{-3mm}
\end{table}

\subsection{Evaluation protocol}
\label{sec:eval_proc}
A standard linear-probe protocol \cite{chen2020simple, lubana2022orchestra} is utilized to evaluate the pre-trained SSL models. In this protocol, the pre-trained SSL model is frozen, and a linear classifier is learned on top of it. In addition to fine-tuning on the whole training set, we also conduct semi-supervised learning with limited labeled data (1\% and 10\%). Following the setup in \cite{tamkin2021viewmaker}, we set the batch size to 128, momentum to 0.9, and the learning rate to 0.01, which is decayed by a factor of 0.1 after 60 and 80 epochs. We perform the training for 100 epochs and report the test results for the last epoch.

\section{Experimental results and discussion}


\subsection{Ablation studies}
\label{sec:ab_study}
\begin{figure}[t]
\centering
\begin{subfigure}{0.333\columnwidth}
    \includegraphics[scale=0.6]{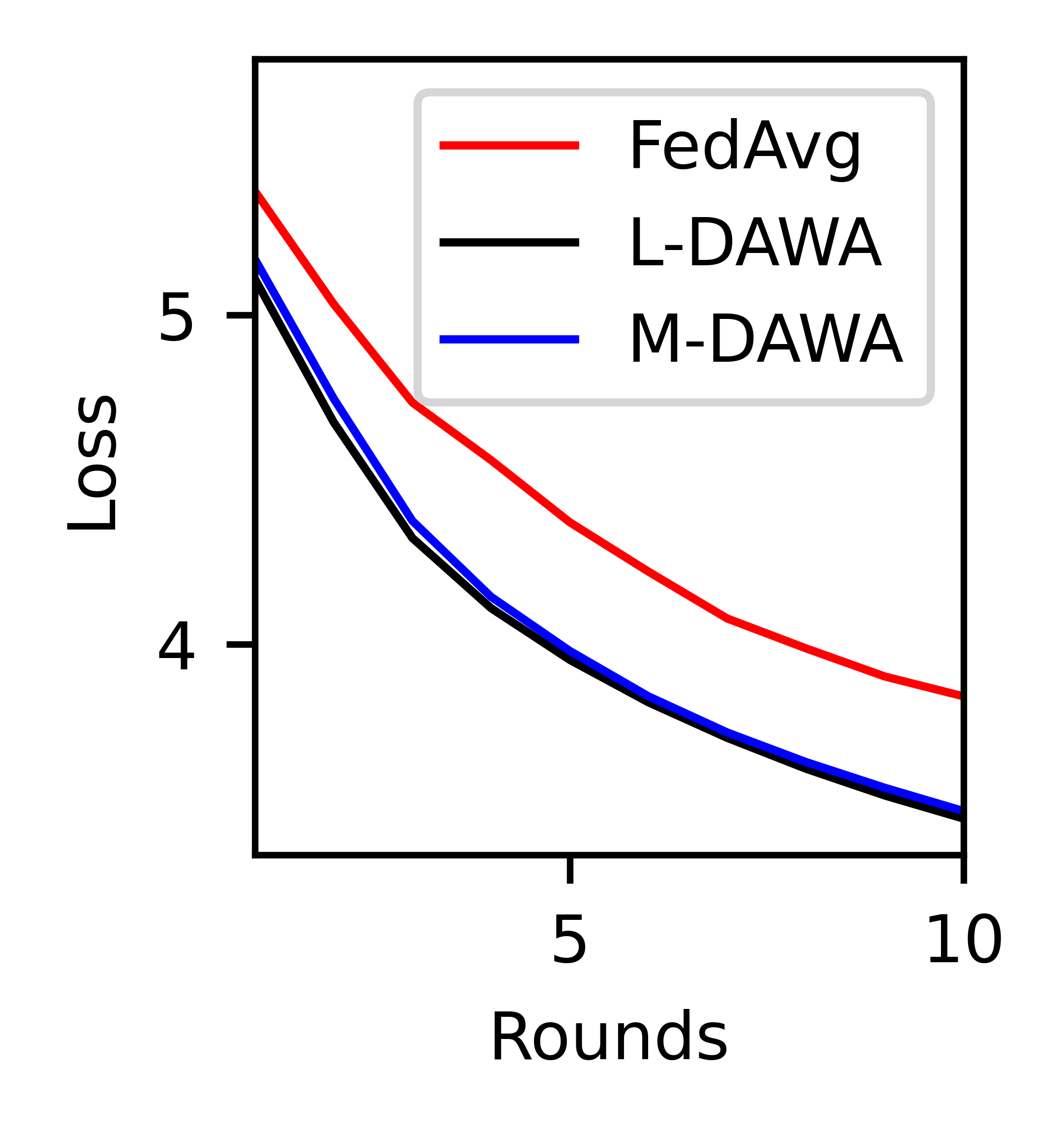}
    \caption{E=1}
\end{subfigure}%
\begin{subfigure}{0.333\columnwidth}
    \includegraphics[scale=0.6]{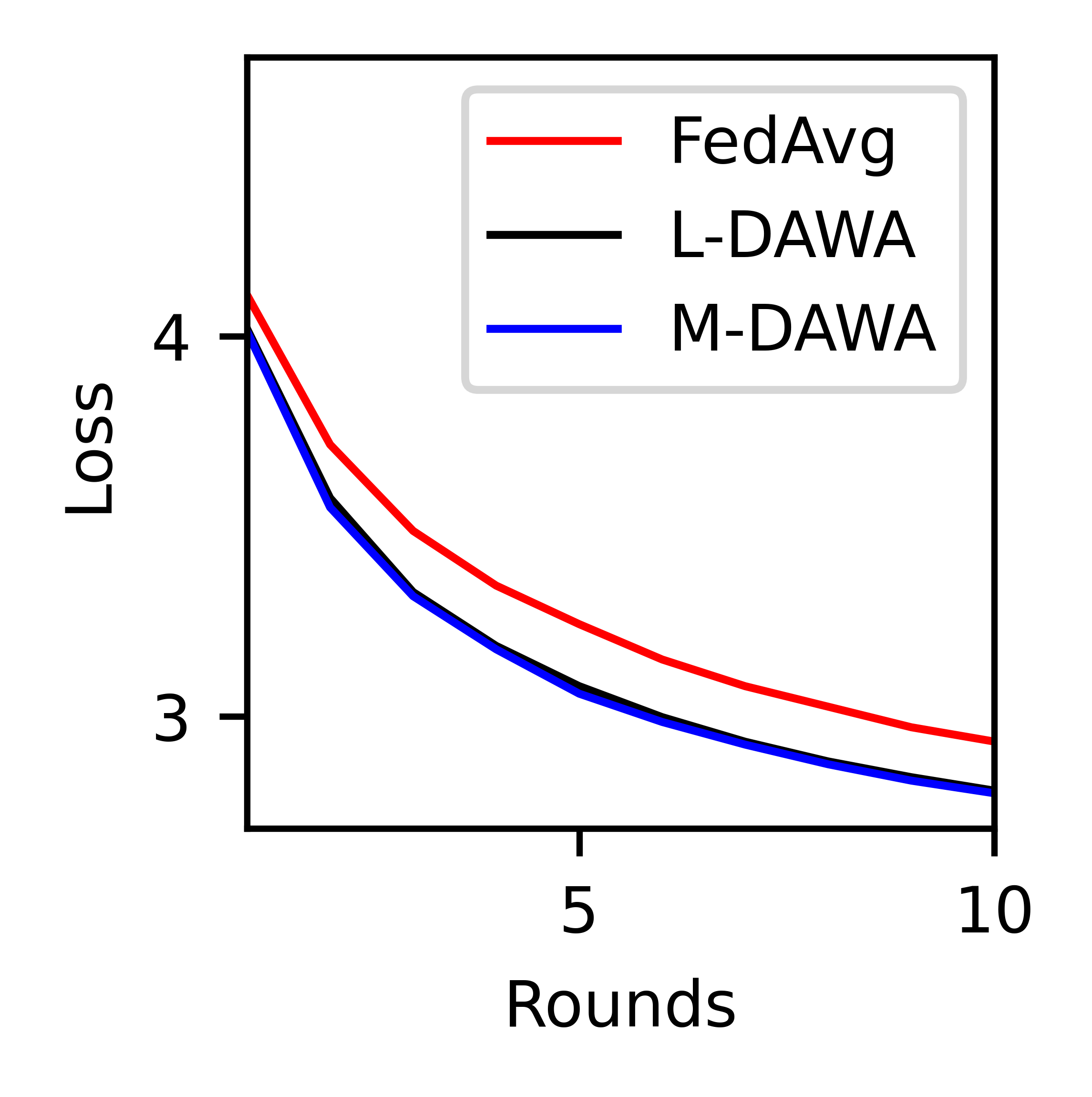}
    \caption{E=5}
\end{subfigure}%
\begin{subfigure}{0.333\columnwidth}
    \includegraphics[scale=0.6]{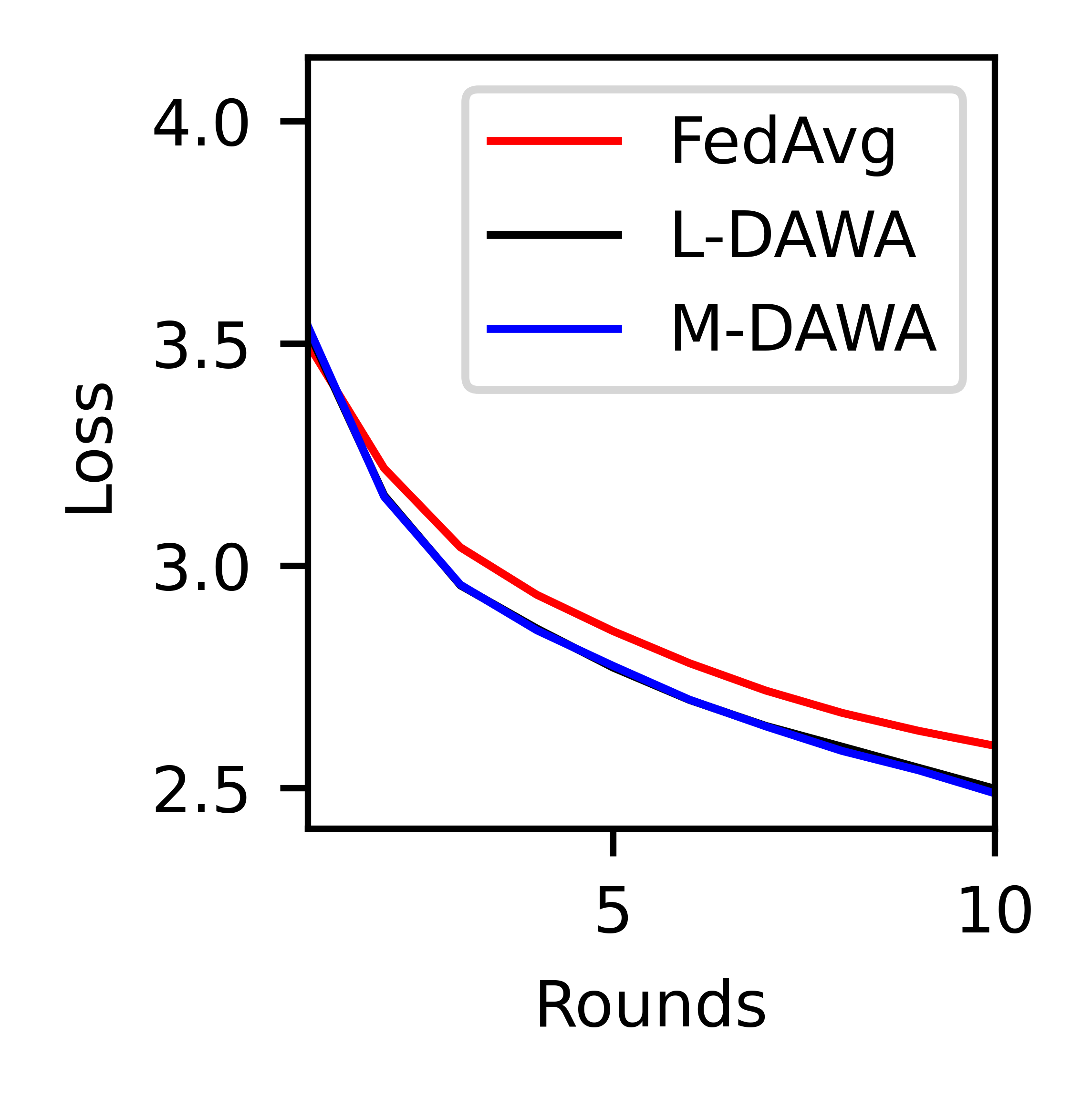}
    \caption{E=10}
\end{subfigure}
    \caption{\small Average local loss of SimCLR for FedAvg, M-DAWA, and L-DAWA: Each method is pre-trained with SimCLR on the Non-iid version of CIFAR-10 for R=10 rounds under the cross-silo (K=10) setting. }
    \label{fig:loss_curves}
    \vspace{-3mm}
\end{figure}

We first perform experiments by comparing L-DAWA and M-DAWA against the baseline FedAvg under the \textit{cross-silo} settings with Non-iid data using the SimCLR method and training them for 10 communication rounds. L-DAWA and M-DAWA consistently achieve higher performance (Table \ref{tab:DAWA-P-vs-DAWA-C}) and converge faster (Figure \ref{fig:loss_curves}) compared to FedAvg. Particularly, L-DAWA obtains the best results on E=1 and E=10 settings. On the other hand, we find that M-DAWA takes 22.65 seconds to complete the aggregation process, which is around 60$\times$ and 78$\times$ slower than L-DAWA and FedAvg, respectively. This computational delay mainly results from the divergence calculation of the entire model.
Therefore, we stick to L-DAWA for all of the rest experiments due to its stable performance and lower resource consumption compared to M-DAWA.

\begin{figure}[t]
\centering
    \includegraphics[width=\linewidth]{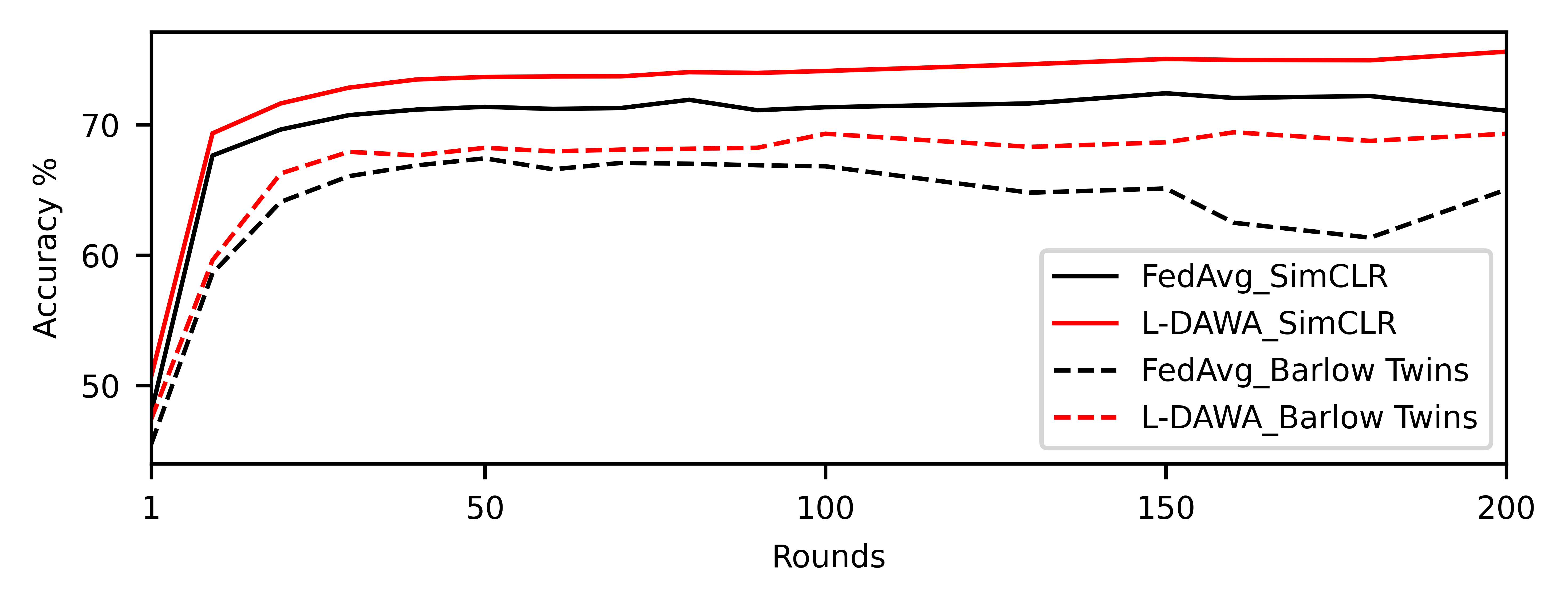}
\caption{\small Variation of linear-probe accuracy (\%) with communication rounds for SimCLR and Barlow Twins. L-DAWA shows stable and improved convergence compared to FedAvg.}
\label{fig:comm_eff}
\vspace{-4mm}
\end{figure}

\begin{table*}[t]
    \centering
    \resizebox{2\columnwidth}{!}{
    \begin{tabular}{l ccc|ccc|ccc|ccc}
         \toprule
         \multirow{3}{*}{Method}& \multicolumn{6}{c}{CIFAR-10} & \multicolumn{6}{c}{CIFAR-100}\\
         \cline{2-13}
         &  \multicolumn{3}{c}{SimCLR} & \multicolumn{3}{c}{Barlow Twins} & \multicolumn{3}{c}{SimCLR} & \multicolumn{3}{c}{Barlow Twins}  \\
        \cline{2-13}
        & \multicolumn{1}{c}{100\%} & 1\% & 10\% & 100\%&  1\% & 10\%  & 100\% &   1\% & 10\%   & 100\% &   1\% & 10\% \\
        \cline{2-13}
         \hline
         FedAvg \cite{mcmahan2017communication}  & 71.07 & 53.32 & 66.13 &  65.02 & 43.29 & 57.06 & 43.85 & 21.35 & 36.49 & 35.70  & 12.93 & 27.04 \\
         Loss \cite{gao2022end}  & 71.34 & 60.14 & 69.91 & 57.12 & 42.70 & 54.10 & 44.69 & 20.90 & 37.14 & 34.76  & 11.08 & 23.88\\
         FedU \cite{zhuang2021collaborative}  & 70.36 & 61.74 & 69.77 & 64.55 & 50.95 & 61.96 & 44.31 & 20.86 & 36.76 & 35.25  & 12.57 & 26.72\\
         EUC \cite{lee2021layer} & 70.51 & 60.77 & 69.37 & 63.60 & 49.62 & 62.61 & 44.89 & 21.96 & 36.90 & 35.44 & 12.43 & 27.07 \\ 
         \hline
          L-DAWA   & 75.60 & 62.19 & 71.40 & 69.31  & 57.62 & \textbf{68.38} & 49.88 & 21.53 & 39.41 & 41.85 &  16.92 & 32.12  \\
            L-DAWA$_{FedAvg}$  & 75.72 & 63.74 & \textbf{73.29} & 69.14 &\textbf{ 58.98} & 67.61 & 49.99 & \textbf{21.87} & 39.57 & 41.49  & \textbf{17.35} & \textbf{32.20} \\
             L-DAWA$_{Loss}$  &\textbf{ 76.55} & 63.86 & 72.82 & 69.46 & 58.74 & 66.82 & 50.29 & 21.43 & 39.72 & \textbf{41.89} & 16.36 & 31.88  \\
             L-DAWA$_{FedU}$  & 76.23 & \textbf{64.06} & 72.89 & \textbf{69.50} & 58.67 & 67.42 &  \textbf{50.59} & 21.72 & \textbf{39.76} & 41.72  & 17.28 & 31.91\\
         
         \bottomrule
          \end{tabular}
    }
    \caption{\small Comparison of L-DAWA with SOTA methods on the Non-iid version ($\alpha$=0.1) of CIFAR-10 and CIFAR-100 under \textit{cross-silo (K=10)} settings. The pre-trained SSL models are separately fine-tuned on 100\%, 10\%, and 1\% data samples of the training set.}
    \vspace{-3mm}
    \label{tab:comp_sota_cross-silo-a}
\end{table*}

\subsection{Communication efficiency}
\label{sec:com_eff}
In this experiment, we analyze the computational efficiency and model convergence behavior of pre-training SimCLR and Barlow Twins in FL with FedAvg and L-DAWA. Figure \ref{fig:comm_eff} shows that L-DAWA offers a faster convergence and requires less number of communication rounds to achieve a given target accuracy for both SimCLR and Barlow Twins compared to FedAvg. Additionally, L-DAWA offers an improved and stable performance for both SSL methods. In contrast, FedAvg requires more rounds to reach a given target accuracy for SimCLR and Barlow Twins. This is evident from the corresponding optimization trajectories for SimCLR and Barlow Twins as shown in Figure \ref{fig:Optimization-a}. 
One can see from Figure \ref{fig:Optimization-a} that the optimization trajectory of FL pre-training of SimCLR with FedAvg lags behind SimCLR with L-DAWA. Whereas the optimization trajectory of FL pre-training of Barlow Twins with FedAvg oscillates more near the optimization point compared to the one with L-DAWA, which also provides a possible reason for the deterioration of the performance near the end of FL pre-training (Figure \ref{fig:comm_eff}).


\subsection{Fairness analysis}
\label{sec:fair_ans}
In this section, we provide a fairness analysis of our proposed method regarding the number of local data samples on the clients. To offer a formal illustration, we conduct two Non-iid federated datasets with 10 clients using CIFAR-10 at a similar level of difficulty for SSL: (1) each client contains the same number of samples with only one single class (W/Sc) without overlap between the classes; (2) the Dirichlet coefficient $\alpha$ is set to $0.1$ with an uneven distribution of the samples among the clients. Table \ref{tab:fairness} shows the linear-probe performance comparison of L-DAWA with FedAvg on these two settings. One can see that FedAvg performs worse on the downstream task under W/Sc setting compared to the case when the clients contain an unequal number of samples Dir(\(\alpha=0.1\)). This is mainly because FedAvg is biased toward the clients with more samples than others (e.g., $\alpha$=0.1 setting). The model optimized with FedAvg would deviate to the dominant clients while the contribution from other clients would be diminished \cite{li2019fair}. This causes the model converges in a local optimum with lower performance on downstream tasks.

Compared to FedAvg, L-DAWA offers higher and more fair performance in these two Non-iid settings, especially for SimCLR. One can observe that L-DAWA with E=5 produces nearly the same results with both configurations. This demonstrates that our proposed method is agnostic to the number of data samples on each client. L-DAWA provides a more balanced optimization over the clients by fairly aggregating the local model weights. Indeed, this would help the model move towards global optima during FL training.
\begin{table}[t]
    \centering
    \resizebox{\columnwidth}{!}{
    \begin{tabular}{llcccccc}
        \hline
        SSL Type & Aggregation Type &  \multicolumn{3}{c}{Non-iid (W/Sc)} & \multicolumn{3}{c}{Non-iid ($\alpha$=0.1)}  \\
         \cline{1-8}
                & & E1 & E5 & E10  & E1 &  E5 & E10 \\
        \cline{3-8}
        \multirow{3}{*}{SimCLR} & FedAvg  & 49.06 &  60.06 & 66.13 &  50.93 & 65.42 & 71.36 \\
         & L-DAWA   & \textbf{60.45 }& \textbf{69.08} & \textbf{73.49} &  \textbf{60.29} & \textbf{70.65} & \textbf{75.60}\\
        \hline
        \multirow{3}{*}{Barlow Twins}& FedAvg & 49.47  & 57.14 & 60.51 & 51.65  & 58.84 & 65.02 \\
        & L-DAWA & \textbf{50.54 } &\textbf{63.46 }& \textbf{65.70}   & \textbf{54.84} & \textbf{65.07} & \textbf{69.31}\\ 
        \hline
    \end{tabular}
    }
    \caption{\small Fairness Evaluation of FedAvg and L-DAWA for SimCLR and Barlow Twins. W/Sc means single class per client. Each method is pre-trained on the Non-iid version of CIFAR-10 for R=200 rounds under the \textit{cross-silo (K=10)} settings.}
    \vspace{-3mm}
    \label{tab:fairness}
\end{table}

\subsection{L-DAWA in linear fine-tuning}
\label{L-SS-E}
In this section, we compare the performance of our proposed methods against the state-of-the-art aggregation strategies, \textit{viz}., FedAvg\cite{mcmahan2017communication}, Loss \cite{gao2022end}, FedU \cite{zhuang2021collaborative} and EUC \cite{lee2021layer} with supervised and semi-supervised fine-tuning schemes under \textit{cross-silo} and \textit{cross-device} settings.

\subsubsection{Cross-silo performance}
In Table~\ref{tab:comp_sota_cross-silo-a}, we compare the linear-probe accuracy of our proposed L-DAWA-related methods against state-of-the-art approaches under \textit{cross-silo (K=10)} setting, fine-tuned with supervised (100\% training set) and semi-supervised (partial training set) schemes  on the datasets of CIFAR-10/100.

First, our methods achieve new SOTA results in all settings. Notably, it gains $5.21$\% and $6.19$\% improvement with $100$\% data fine-tuning on CIFAR-10/100 datasets, respectively. As for the more difficult semi-supervised evaluation, the SOTA accuracy is increased by $2.32$\% ($1$\% data setting), $3.38$\% ($10$\% data setting) and $0.52$\% ($1$\% data setting), $2.62$\% ($10$\% data setting) on CIFAR-10/100, respectively.

Second, L-DAWA-related approaches obtain similar improvements for both SimCLR and Barlow Twins. This indicates that our methods are agnostic to contrastive or non-contrastive SSL models.

Third, L-DAWA achieves higher performance on all of the settings compared to the baselines (FedAvg, Loss, FedU, and EUC). This highlights the importance of integrating divergence into aggregation during FL training. The angular divergence plays a similar role of momentum, accelerating convergence toward the global optima in the relevant direction while diminishing oscillations during FL optimization. In contrast, FedAvg and Loss-based aggregation would be dramatically influenced by the dominant clients with larger amounts of data or lower loss, leading to a deviation of the optimization trajectory.

Interestingly, the modified versions of FedAvg, Loss, and FedU (L-DAWA$_{FedAvg}$, L-DAWA$_{Loss}$ and L-DAWA$_{FedU}$) provide better and unbiased results once the layer-wise divergence is introduced into the aggregation. This is mainly because the integration of angular divergence offers a corrective effect with respect to the original optimization trajectory (Figure \ref{fig:Optimization-a}). Indeed, the combination of L-DAWA with the existing SOTA methods boosts the performance at a large scale and provides various candidates for different settings. For instance, L-DAWA$_{Loss}$ and L-DAWA$_{FedU}$ achieve the highest accuracy in $100$\% data setting, while L-DAWA$_{FedAvg}$ performs best in the most challenging setup ($1$\% semi-supervised evaluation).

We further evaluate L-DAWA on a larger dataset Tiny ImageNet, compared with SOTA approaches (Table \ref{tab:tiny}). A large margin of improvements ($6.25$\%, $3.50$\%, $4.35$\%) is obtained by L-DAWA for the settings of 1/5/10 local epochs, respectively. This demonstrates that our proposed method has a good generalization on a large dataset.
 
We also show the results of our proposed method on different Non-iid levels of \textit{cross-silo} by scaling Dirichlet coefficient $\alpha$ from $0.1$ to $0.6$ (Table \ref{tab:diff_alpha}). One can see that the performance of all methods shows a slight increase trend with Non-iid levels decreasing. Noticeably, L-DAWA obtains higher performance at all levels of Non-iid settings, especially for SimCLR ($\alpha=0.1$) with $9.37$\% improvement. Interestingly, the results on the $\alpha=0.2$ setting are better than others. This indicates that the Non-iid level for image-SSL may be partially determined by actual class labels.

\begin{table}[!]
    \centering
    
    \scalebox{0.85}{
    \begin{tabular}{llcccc}
    \toprule
          & & \multicolumn{4}{c}{$\alpha$ values} \\
        \cline{3-6}
         Non-iid levels & & 0.1 & 0.2 & 0.4 & 0.6   \\
        \hline
        \multirow{3}{*}{SimCLR} & FedAvg & 50.92 & 53.26 & 53.98 & 53.94   \\
          & L-DAWA & \textbf{60.29} & \textbf{63.41} & \textbf{62.98} & \textbf{63.40}  \\
        \hline
        \multirow{3}{*}{Barlow Twins} & FedAvg & 53.45 & 54.72 & 54.18 & 54.82  \\
          & L-DAWA & \textbf{54.84} & \textbf{57.90} & \textbf{54.61} & \textbf{57.89}   \\
    \bottomrule
    \end{tabular}
    }
    \caption{\small Performance of FedAvg and L-DAWA for SimCLR and Barlow Twins within different Non-iid levels ($\alpha$ values). Each method is pre-trained on CIFAR-10 with 1 local epoch for R=200 rounds under the \textit{cross-silo (K=10)} settings.}
    \vspace{-3mm}
    \label{tab:diff_alpha}
\end{table}

\subsubsection{Cross-device performance}
Compared to \textit{cross-silo}, \textit{cross-device} setting is more challenging due to its nature of heterogeneous data distribution. One can observe from Table \ref{tab:comp_sota_cross-device-a} that our proposed methods still perform better than all baselines with a slight improvement of $1.26$\%/$0.25$\% (SimCLR) and $0.25$\%/$0.85$\% (Barlow Twins) on CIFAR-10/100, respectively. Also, the combination of L-DAWA with the existing methods (L-DAWA$_{FedAvg}$ and L-DAWA$_{Loss}$) provide the best performance at most of the settings, suggesting that in \textit{cross-device} settings the integration of angular divergence is necessary to boost performance.


\begin{table}[t]
    \centering
    
    \scalebox{0.85}{
    \begin{tabular}{lccc}
    \toprule
         Aggregation Type & E1 & E5 & E10 \\
         \hline
             FedAvg  & 16.87 & 27.70 & 32.92 \\
             Loss  & 15.25 & 28.47 & 33.37 \\
             FedU  & 16.41 & 27.46 & 32.63  \\
        \hline
             L-DAWA  & \textbf{23.12} &\textbf{ 31.97}&\textbf{37.72} \\
    \bottomrule
    \end{tabular}
    }
    \caption{\small Evaluation with SimCLR on the Non-iid version of Tiny ImageNet under the \textit{cross-silo (K=10)} setting.}
    \vspace{-5mm}
    \label{tab:tiny}
\end{table}

\begin{table}[!]
    \centering
    
    \resizebox{\columnwidth}{!}{
    \begin{tabular}{l cccc}
         \toprule
         Method & \multicolumn{2}{c}{CIFAR-10} & \multicolumn{2}{c}{CIFAR-100}\\
        \hline
         &  SimCLR & Barlow Twins & SimCLR & Barlow Twins\\
        \cline{2-5}
         FedAvg   & 68.66 & 62.07 &  44.59 & 32.65 \\
          Loss   & 66.09 & 56.40  & 44.83 &  33.27  \\
          FedU    & 68.52 &  61.43 & 44.56  & 32.89 \\
        \hline
         L-DAWA  & 68.20 & 58.25 & 45.04  &  \textbf{34.12}   \\
         L-DAWA$_{FedAvg}$  & \textbf{69.92} &  \textbf{62.32} & 44.19  & 33.20 \\
         L-DAWA$_{Loss}$ & 68.79 &  61.36 &  \textbf{45.08}  & 31.93  \\
         L-DAWA$_{FedU}$  & 69.69 & 62.19 & 44.97 &32.84 \\
         \bottomrule
    \end{tabular}
    }
    \caption{\small Comparison of the proposed aggregation strategy with state-of-the-art methods on the Non-iid version ($\alpha$=0.1) of CIFAR-10 and CIFAR-100 under \textit{cross-device (K=100)} setting.} 
    \label{tab:comp_sota_cross-device-a}

\end{table}


        

\subsection{Evaluation on transfer learning} 
\label{sec:trans_learning}

We further evaluate the generalization of the learned features from FL pre-training by fine-tuning the resulting model on a different dataset. Such evaluation helps in assessing whether the pre-trained representations can be transferred to different downstream tasks. We follow the same procedure that is adopted for linear evaluation. Specifically, we first perform FL pre-training on Tiny ImageNet followed by linear-probe evaluation on CIFAR10/100. 

One can see from Table \ref{tab:Transfer_Learning} that L-DAWA generalizes well for both CIFAR-10/100 compared to other aggregation strategies in the \textit{cross-silo} settings. Particularly, it obtains $4.4$\% and $5.7$\% improvements under these two cross-dataset settings, respectively.
Additionally, with the integration of angular divergence into FedAvg, Loss, and FedU, a significant performance boost is obtained for these methods.

\begin{table}[!]
    \centering
    \scalebox{0.85}{
    \begin{tabular}{l cc}
    \toprule

        & Tiny ImageNet & Tiny ImageNet \\
        Methods & $\rightarrow$ CIFAR-10 & $\rightarrow$ CIFAR-100 \\
        
          \cline{1-3}
         \cline{1-3}
         FedAvg & 77.46  & 52.11  \\
         Loss & 77.52  & 52.68 \\
         FedU &  76.70 &52.25  \\
         \hline
         L-DAWA &  \textbf{81.87}  & \textbf{57.81} \\
         L-DAWA-W$_{FedAvg}$ & 81.61 & 57.75 \\
         L-DAWA-W$_{Loss}$ & 81.87  &57.86  \\
         L-DAWA-W$_{FedU}$ & 81.62  & 57.34 \\
    \bottomrule
    \end{tabular}
    }
    \caption{\small Comparison of the proposed aggregation strategy with state-of-the-art methods for transfer learning with cross-dataset evaluation under \textit{cross-silo (K=10)} setting.}
    \vspace{-5mm}
    \label{tab:Transfer_Learning}
\end{table}

\section{Conclusion}
In this paper, we proposed layer-wise divergence aware weight aggregation (L-DAWA) for SSL pre-training in FL. We empirically show that the SOTA methods get biased towards the clients' metadata (number of samples and loss). To reduce such bias, L-DAWA scales the weighting of each layer of clients' models, based on the measure of  \textit{layer-wise} \textit{angular divergence} with previous global model. Extensive experiments show that L-DAWA obtained a new SOTA performance in the \textit{cross-silo} and \textit{cross-device} settings with both contrastive and non-contrastive SSL methods.

{\small
\bibliographystyle{ieee_fullname}
\bibliography{egbib}

\begin{thebibliography}{10}\itemsep=-1pt

\bibitem{Benaim_2020_CVPR}
Sagie Benaim, Ariel Ephrat, Oran Lang, Inbar Mosseri, William~T. Freeman,
  Michael Rubinstein, Michal Irani, and Tali Dekel.
\newblock Speednet: Learning the speediness in videos.
\newblock In {\em Proceedings of the IEEE/CVF Conference on Computer Vision and
  Pattern Recognition (CVPR)}, June 2020.

\bibitem{beutel2020flower}
Daniel~J Beutel, Taner Topal, Akhil Mathur, Xinchi Qiu, Titouan Parcollet, and
  Nicholas~D Lane.
\newblock Flower: A friendly federated learning research framework.
\newblock {\em arXiv preprint arXiv:2007.14390}, 2020.

\bibitem{caldarola2022improving}
Debora Caldarola, Barbara Caputo, and Marco Ciccone.
\newblock Improving generalization in federated learning by seeking flat
  minima.
\newblock In {\em European Conference on Computer Vision}, pages 654--672.
  Springer, 2022.

\bibitem{caron2020unsupervised}
Mathilde Caron, Ishan Misra, Julien Mairal, Priya Goyal, Piotr Bojanowski, and
  Armand Joulin.
\newblock Unsupervised learning of visual features by contrasting cluster
  assignments.
\newblock {\em Advances in Neural Information Processing Systems},
  33:9912--9924, 2020.

\bibitem{chen2020simple}
Ting Chen, Simon Kornblith, Mohammad Norouzi, and Geoffrey Hinton.
\newblock A simple framework for contrastive learning of visual
  representations.
\newblock In {\em International conference on machine learning}, pages
  1597--1607. PMLR, 2020.

\bibitem{chen2021exploring}
Xinlei Chen and Kaiming He.
\newblock Exploring simple siamese representation learning.
\newblock In {\em Proceedings of the IEEE/CVF Conference on Computer Vision and
  Pattern Recognition}, pages 15750--15758, 2021.

\bibitem{dong2022spherefed}
Xin Dong, Sai~Qian Zhang, Ang Li, and HT Kung.
\newblock Spherefed: Hyperspherical federated learning.
\newblock {\em arXiv preprint arXiv:2207.09413}, 2022.

\bibitem{duan2021flexible}
Moming Duan, Duo Liu, Xinyuan Ji, Yu Wu, Liang Liang, Xianzhang Chen, Yujuan
  Tan, and Ao Ren.
\newblock Flexible clustered federated learning for client-level data
  distribution shift.
\newblock {\em IEEE Transactions on Parallel and Distributed Systems}, 2021.

\bibitem{falcon2019pytorch}
William Falcon et~al.
\newblock Pytorch lightning.
\newblock {\em GitHub. Note: https://github.
  com/PyTorchLightning/pytorch-lightning}, 3(6), 2019.

\bibitem{gao2022end}
Yan Gao, Titouan Parcollet, Salah Zaiem, Javier Fernandez-Marques, Pedro~PB de
  Gusmao, Daniel~J Beutel, and Nicholas~D Lane.
\newblock End-to-end speech recognition from federated acoustic models.
\newblock In {\em ICASSP 2022-2022 IEEE International Conference on Acoustics,
  Speech and Signal Processing (ICASSP)}, pages 7227--7231. IEEE, 2022.

\bibitem{goyal2022vision}
Priya Goyal, Quentin Duval, Isaac Seessel, Mathilde Caron, Mannat Singh, Ishan
  Misra, Levent Sagun, Armand Joulin, and Piotr Bojanowski.
\newblock Vision models are more robust and fair when pretrained on uncurated
  images without supervision.
\newblock {\em arXiv preprint arXiv:2202.08360}, 2022.

\bibitem{grill2020bootstrap}
Jean-Bastien Grill, Florian Strub, Florent Altch{\'e}, Corentin Tallec, Pierre
  Richemond, Elena Buchatskaya, Carl Doersch, Bernardo Avila~Pires, Zhaohan
  Guo, Mohammad Gheshlaghi~Azar, et~al.
\newblock Bootstrap your own latent-a new approach to self-supervised learning.
\newblock {\em Advances in neural information processing systems},
  33:21271--21284, 2020.

\bibitem{he2020momentum}
Kaiming He, Haoqi Fan, Yuxin Wu, Saining Xie, and Ross Girshick.
\newblock Momentum contrast for unsupervised visual representation learning.
\newblock In {\em Proceedings of the IEEE/CVF conference on computer vision and
  pattern recognition}, pages 9729--9738, 2020.

\bibitem{he2016deep}
Kaiming He, Xiangyu Zhang, Shaoqing Ren, and Jian Sun.
\newblock Deep residual learning for image recognition.
\newblock In {\em Proceedings of the IEEE conference on computer vision and
  pattern recognition}, pages 770--778, 2016.

\bibitem{jain2021biometrics}
Anil~K Jain, Debayan Deb, and Joshua~J Engelsma.
\newblock Biometrics: Trust, but verify.
\newblock {\em arXiv preprint arXiv:2105.06625}, 2021.

\bibitem{kairouz2021advances}
Peter Kairouz, H~Brendan McMahan, Brendan Avent, Aur{\'e}lien Bellet, Mehdi
  Bennis, Arjun~Nitin Bhagoji, Kallista Bonawitz, Zachary Charles, Graham
  Cormode, Rachel Cummings, et~al.
\newblock Advances and open problems in federated learning.
\newblock {\em Foundations and Trends{\textregistered} in Machine Learning},
  14(1--2):1--210, 2021.

\bibitem{komodakis2018unsupervised}
Nikos Komodakis and Spyros Gidaris.
\newblock Unsupervised representation learning by predicting image rotations.
\newblock In {\em International Conference on Learning Representations (ICLR)},
  2018.

\bibitem{krizhevsky2009learning}
Alex Krizhevsky, Geoffrey Hinton, et~al.
\newblock Learning multiple layers of features from tiny images.
\newblock 2009.

\bibitem{lee2021layer}
Sunwoo Lee, Tuo Zhang, Chaoyang He, and Salman Avestimehr.
\newblock Layer-wise adaptive model aggregation for scalable federated
  learning.
\newblock {\em arXiv preprint arXiv:2110.10302}, 2021.

\bibitem{li2020review}
Li Li, Yuxi Fan, Mike Tse, and Kuo-Yi Lin.
\newblock A review of applications in federated learning.
\newblock {\em Computers \& Industrial Engineering}, 149:106854, 2020.

\bibitem{li2021model}
Qinbin Li, Bingsheng He, and Dawn Song.
\newblock Model-contrastive federated learning.
\newblock In {\em Proceedings of the IEEE/CVF Conference on Computer Vision and
  Pattern Recognition}, pages 10713--10722, 2021.

\bibitem{li2020federated}
Tian Li, Anit~Kumar Sahu, Manzil Zaheer, Maziar Sanjabi, Ameet Talwalkar, and
  Virginia Smith.
\newblock Federated optimization in heterogeneous networks.
\newblock {\em Proceedings of Machine learning and systems}, 2:429--450, 2020.

\bibitem{li2019fair}
Tian Li, Maziar Sanjabi, Ahmad Beirami, and Virginia Smith.
\newblock Fair resource allocation in federated learning.
\newblock {\em arXiv preprint arXiv:1905.10497}, 2019.

\bibitem{li2022understanding}
Ziwei Li, Hong-You Chen, Han~Wei Shen, and Wei-Lun Chao.
\newblock Understanding federated learning through loss landscape
  visualizations: A pilot study.
\newblock In {\em Workshop on Federated Learning: Recent Advances and New
  Challenges (in Conjunction with NeurIPS 2022)}.

\bibitem{liu2020accelerating}
Wei Liu, Li Chen, Yunfei Chen, and Wenyi Zhang.
\newblock Accelerating federated learning via momentum gradient descent.
\newblock {\em IEEE Transactions on Parallel and Distributed Systems},
  31(8):1754--1766, 2020.

\bibitem{lubana2022orchestra}
Ekdeep~Singh Lubana, Chi~Ian Tang, Fahim Kawsar, Robert~P Dick, and Akhil
  Mathur.
\newblock Orchestra: Unsupervised federated learning via globally consistent
  clustering.
\newblock {\em arXiv preprint arXiv:2205.11506}, 2022.

\bibitem{Ma_2022_CVPR}
Xiaosong Ma, Jie Zhang, Song Guo, and Wenchao Xu.
\newblock Layer-wised model aggregation for personalized federated learning.
\newblock In {\em Proceedings of the IEEE/CVF Conference on Computer Vision and
  Pattern Recognition (CVPR)}, pages 10092--10101, June 2022.

\bibitem{mcmahan2017communication}
Brendan McMahan, Eider Moore, Daniel Ramage, Seth Hampson, and Blaise~Aguera y
  Arcas.
\newblock Communication-efficient learning of deep networks from decentralized
  data.
\newblock In {\em Artificial intelligence and statistics}, pages 1273--1282.
  PMLR, 2017.

\bibitem{Mendieta_2022_CVPR}
Matias Mendieta, Taojiannan Yang, Pu Wang, Minwoo Lee, Zhengming Ding, and Chen
  Chen.
\newblock Local learning matters: Rethinking data heterogeneity in federated
  learning.
\newblock In {\em Proceedings of the IEEE/CVF Conference on Computer Vision and
  Pattern Recognition (CVPR)}, pages 8397--8406, June 2022.

\bibitem{Menon_2020_CVPR}
Sachit Menon, Alexandru Damian, Shijia Hu, Nikhil Ravi, and Cynthia Rudin.
\newblock Pulse: Self-supervised photo upsampling via latent space exploration
  of generative models.
\newblock In {\em Proceedings of the IEEE/CVF Conference on Computer Vision and
  Pattern Recognition (CVPR)}, June 2020.

\bibitem{noroozi2016unsupervised}
Mehdi Noroozi and Paolo Favaro.
\newblock Unsupervised learning of visual representations by solving jigsaw
  puzzles.
\newblock In {\em European conference on computer vision}, pages 69--84.
  Springer, 2016.

\bibitem{ou2020lincos}
Wei-Feng Ou, Lai-Man Po, Chang Zhou, Yu-Jia Zhang, Li-Tong Feng, Yasar Abbas~Ur
  Rehman, and Yu-Zhi Zhao.
\newblock Lincos-softmax: Learning angle-discriminative face representations
  with linearity-enhanced cosine logits.
\newblock {\em IEEE Access}, 8:109758--109769, 2020.

\bibitem{park2020sinet}
Hyojin Park, Lars Sjosund, YoungJoon Yoo, Nicolas Monet, Jihwan Bang, and Nojun
  Kwak.
\newblock Sinet: Extreme lightweight portrait segmentation networks with
  spatial squeeze module and information blocking decoder.
\newblock In {\em Proceedings of the IEEE/CVF Winter Conference on Applications
  of Computer Vision}, pages 2066--2074, 2020.

\bibitem{pathak2016context}
Deepak Pathak, Philipp Krahenbuhl, Jeff Donahue, Trevor Darrell, and Alexei~A
  Efros.
\newblock Context encoders: Feature learning by inpainting.
\newblock In {\em Proceedings of the IEEE conference on computer vision and
  pattern recognition}, pages 2536--2544, 2016.

\bibitem{reddi2020adaptive}
Sashank Reddi, Zachary Charles, Manzil Zaheer, Zachary Garrett, Keith Rush,
  Jakub Kone{\v{c}}n{\`y}, Sanjiv Kumar, and H~Brendan McMahan.
\newblock Adaptive federated optimization.
\newblock {\em arXiv preprint arXiv:2003.00295}, 2020.

\bibitem{rehman2022federated}
Yasar Abbas~Ur Rehman, Yan Gao, Jiajun Shen, Pedro Porto~Buarque de Gusmao, and
  Nicholas Lane.
\newblock Federated self-supervised learning for video understanding.
\newblock {\em arXiv preprint arXiv:2207.01975}, 2022.

\bibitem{sattler2020clustered}
Felix Sattler, Klaus-Robert M{\"u}ller, and Wojciech Samek.
\newblock Clustered federated learning: Model-agnostic distributed multitask
  optimization under privacy constraints.
\newblock {\em IEEE transactions on neural networks and learning systems},
  32(8):3710--3722, 2020.

\bibitem{schroff2015facenet}
Florian Schroff, Dmitry Kalenichenko, and James Philbin.
\newblock Facenet: A unified embedding for face recognition and clustering.
\newblock In {\em Proceedings of the IEEE conference on computer vision and
  pattern recognition}, pages 815--823, 2015.

\bibitem{tamkin2021viewmaker}
Alex Tamkin, Mike Wu, and Noah Goodman.
\newblock Viewmaker networks: Learning views for unsupervised representation
  learning.
\newblock In {\em International Conference on Learning Representations}, 2021.

\bibitem{wang2021unsupervised}
Guangting Wang, Yizhou Zhou, Chong Luo, Wenxuan Xie, Wenjun Zeng, and Zhiwei
  Xiong.
\newblock Unsupervised visual representation learning by tracking patches in
  video.
\newblock In {\em Proceedings of the IEEE/CVF Conference on Computer Vision and
  Pattern Recognition}, pages 2563--2572, 2021.

\bibitem{wang2022friends}
Heqiang Wang and Jie Xu.
\newblock Friends to help: Saving federated learning from client dropout.
\newblock {\em arXiv preprint arXiv:2205.13222}, 2022.

\bibitem{wortsman2022model}
Mitchell Wortsman, Gabriel Ilharco, Samir~Ya Gadre, Rebecca Roelofs, Raphael
  Gontijo-Lopes, Ari~S Morcos, Hongseok Namkoong, Ali Farhadi, Yair Carmon,
  Simon Kornblith, et~al.
\newblock Model soups: averaging weights of multiple fine-tuned models improves
  accuracy without increasing inference time.
\newblock In {\em International Conference on Machine Learning}, pages
  23965--23998. PMLR, 2022.

\bibitem{xu2019self}
Dejing Xu, Jun Xiao, Zhou Zhao, Jian Shao, Di Xie, and Yueting Zhuang.
\newblock Self-supervised spatiotemporal learning via video clip order
  prediction.
\newblock In {\em Proceedings of the IEEE/CVF Conference on Computer Vision and
  Pattern Recognition}, pages 10334--10343, 2019.

\bibitem{yang2021instance}
Ceyuan Yang, Zhirong Wu, Bolei Zhou, and Stephen Lin.
\newblock Instance localization for self-supervised detection pretraining.
\newblock In {\em Proceedings of the IEEE/CVF Conference on Computer Vision and
  Pattern Recognition}, pages 3987--3996, 2021.

\bibitem{zbontar2021barlow}
Jure Zbontar, Li Jing, Ishan Misra, Yann LeCun, and St{\'e}phane Deny.
\newblock Barlow twins: Self-supervised learning via redundancy reduction.
\newblock In {\em International Conference on Machine Learning}, pages
  12310--12320. PMLR, 2021.

\bibitem{zhang2020federated}
Fengda Zhang, Kun Kuang, Zhaoyang You, Tao Shen, Jun Xiao, Yin Zhang, Chao Wu,
  Yueting Zhuang, and Xiaolin Li.
\newblock Federated unsupervised representation learning.
\newblock {\em arXiv preprint arXiv:2010.08982}, 2020.

\bibitem{zhang2021parameterized}
Jie Zhang, Song Guo, Xiaosong Ma, Haozhao Wang, Wenchao Xu, and Feijie Wu.
\newblock Parameterized knowledge transfer for personalized federated learning.
\newblock {\em Advances in Neural Information Processing Systems},
  34:10092--10104, 2021.

\bibitem{zhang2016colorful}
Richard Zhang, Phillip Isola, and Alexei~A Efros.
\newblock Colorful image colorization.
\newblock In {\em European conference on computer vision}, pages 649--666.
  Springer, 2016.

\bibitem{zhao2018federated}
Yue Zhao, Meng Li, Liangzhen Lai, Naveen Suda, Damon Civin, and Vikas Chandra.
\newblock Federated learning with non-iid data.
\newblock {\em arXiv preprint arXiv:1806.00582}, 2018.

\bibitem{zhuang2021collaborative}
Weiming Zhuang, Xin Gan, Yonggang Wen, Shuai Zhang, and Shuai Yi.
\newblock Collaborative unsupervised visual representation learning from
  decentralized data.
\newblock In {\em Proceedings of the IEEE/CVF International Conference on
  Computer Vision}, pages 4912--4921, 2021.

\bibitem{zhuang2022divergence}
Weiming Zhuang, Yonggang Wen, and Shuai Zhang.
\newblock Divergence-aware federated self-supervised learning.
\newblock {\em arXiv preprint arXiv:2204.04385}, 2022.

\end{thebibliography}
}


\end{document}